\UseRawInputEncoding
\pdfoutput=1
\documentclass[10pt,final,a4paper,oneside,onecolumn]{article}

\edef\reptitle{White paper on\\Selected Environmental Parameters affecting Autonomous Vehicle (AV) Sensors}
\edef\shorttitle{Selected Environmental Parameters affecting AV Sensors}
\def\repauthor{James Lee Wei Shung\\Andrea Piazzoni\\Roshan Vijay\\Lincoln Ang Hon Kin\\Niels de Boer}

\makeatletter
\def\@maketitle{
	\newpage
	\null
	\vspace{-8em}
	\begin{center}
		\begin{tabular}{l c r}
			\multicolumn{3}{c}{\includegraphics[width=0.6\linewidth]{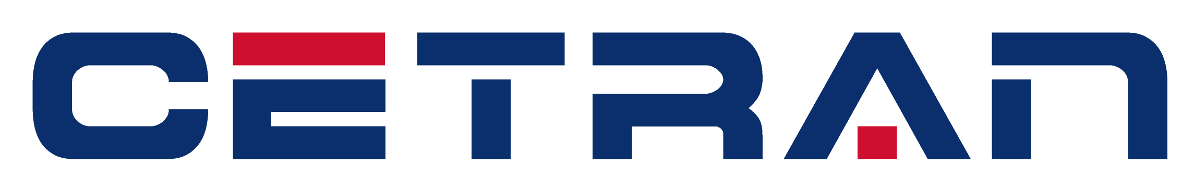}}\vspace{3em} \\
			\includegraphics[width=0.25\linewidth]{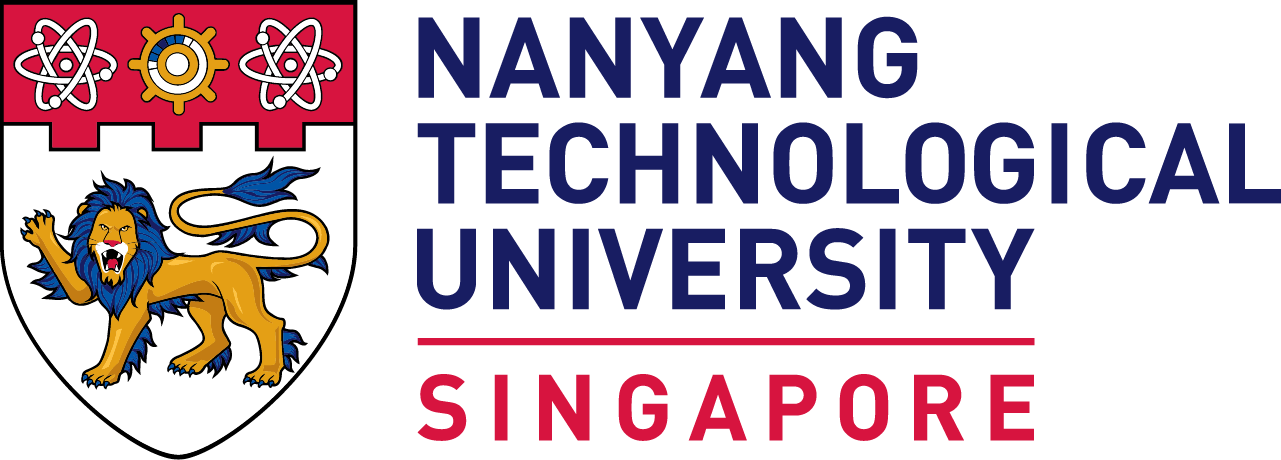} & \hspace{4em} &
			\includegraphics[width=0.3\linewidth]{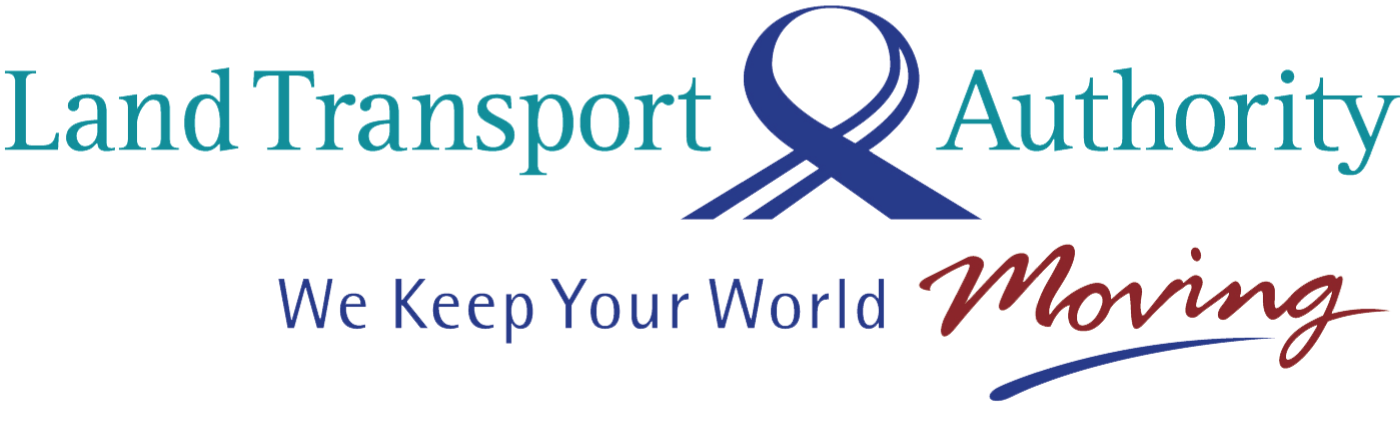}
		\end{tabular}\\
		\vskip 12em
		\let \footnote \thanks
		{\LARGE \@title \par}
		\vskip 1em
		{\large Version 1.0 from September 6, 2023}
		\vskip 1.5em
		{\large
			\lineskip .5em
			\begin{tabular}[t]{c}
				James Lee Wei Shung\\
				Andrea Piazzoni\\
				Roshan Vijay\\
				Lincoln Ang Hon Kin\\
				Niels de Boer
			\end{tabular}\par}
	\end{center}
	\par
	\vskip 1.5em}
\makeatother

\usepackage[utf8]{inputenc}
\usepackage[a4paper,left=2.0cm,right=2.0cm,top=3cm,bottom=3cm]{geometry} 
\usepackage[T1]{fontenc}                        
\usepackage{lmodern}                            
\usepackage[cmex10]{amsmath}                    
\usepackage{amssymb}                            
\usepackage{amsthm}                             
\usepackage{gensymb}
\usepackage{textcomp}
\usepackage{dsfont}                             
\usepackage{mathrsfs}                           
\usepackage[pdftex]{graphicx}                   
\usepackage{array}                              
\usepackage{upgreek}                            
\usepackage[style=ieee,doi=false,isbn=false,url=false,date=year,backend=biber,mincitenames=2,maxcitenames=2,mincitenames=1,uniquelist=false,uniquename=false,giveninits=true,sorting=nty]{biblatex}
\usepackage{stfloats}                           
\usepackage{float}
\usepackage{multirow}                           
\usepackage{subfig}                             
\usepackage{fancyhdr}                           
\usepackage[official,left]{eurosym}             
\usepackage{appendix}                           
\usepackage{xspace}                             
\usepackage{microtype}							
\usepackage{verbatim}                           
\usepackage[dutch,USenglish]{babel}             
\usepackage{wrapfig}                            
\usepackage{longtable}                          
\usepackage{pgfplots}                           
\usepackage{calc}                               
\usepackage{lscape}								
\usepackage[breaklinks=true,hidelinks,          
            bookmarksnumbered=true]{hyperref}   
\usepackage{vhistory}
\usepackage{makecell}
\usepackage[toc]{glossaries}                    
\usepackage[normalem]{ulem}                     
\usepackage{booktabs}							
\usepackage[bottom]{footmisc}                   
\usepackage{csvsimple}  						
\usepackage{makecell}  						
\usepackage{longtable}
\usepackage{acronym}                            
\usepackage{titlesec}
\usepackage{setspace}
\usepackage{changepage}
\usepackage{xcolor}

\usepackage{enumitem}

\usepackage{refcount}

\usepackage{rotating} 

\fancypagestyle{plain}{\fancyhf{}

	\setlength{\headheight}{17pt}
	\addtolength{\footskip}{-5pt}}

\newlength\figurewidth
\newlength\figureheight
\pgfplotsset{every axis/.append style={
		scaled y ticks=false,
		scaled x ticks=false,
		y tick label style={/pgf/number format/fixed},
		x tick label style={/pgf/number format/fixed},
		legend style={font=\small}},
	compat=1.9}                                 
\usetikzlibrary{arrows.meta}                    
\usetikzlibrary{external}                       

\theoremstyle{plain}    
\theoremstyle{definition}     
\theoremstyle{remark}\newtheorem{remarkenv}{Remark}[section]        
                       {\hfill$\lozenge$\end{remarkenv}}            

\graphicspath{{figures/}} 
\fnbelowfloat                                       
\hypersetup{pdftitle={\reptitle},
            pdfauthor={\repauthor},
            colorlinks=true,
            urlcolor=blue,
        	linkcolor=black,
        	citecolor=black}	               


\newlength\ndist                                
\newlength\nheight                              
\newlength\nwidth                                   
\newlength\nsep                                     

\newcommand\disclaimertext{%
	\begin{center} \textbf{Disclaimer} \end{center} 
	\footnotesize
        \begin{center} This white paper was developed with support from the Urban Mobility Grand Challenge Fund\\by the Land Transport Authority of Singapore (No. UMGC-L010).
        \end{center}
}
\newcommand\disclaimernotice{%
	\begin{tikzpicture}[remember picture,overlay]
	\node[anchor=south,yshift=100pt] at (current page.south) {\fbox{\parbox{\dimexpr\textwidth-\fboxsep-\fboxrule\relax}{\disclaimertext}}};
	\end{tikzpicture}%
}

\bibliography{references}

\begin{document}

\title{\Huge\textbf{\reptitle}}

\author{}

\maketitle

\vfill

\clearpage

\setlength\ndist{5em}   
\setlength\nheight{8em} 
\selectlanguage{USenglish}
\pagenumbering{roman}

\vfill

\clearpage

\tableofcontents
\disclaimernotice

\cleardoublepage

\pagenumbering{arabic}

\setcounter{table}{0}

\section{Abstract/Executive summary}
Autonomous Vehicles (AVs) being developed these days rely on various sensor technologies to sense and perceive the world around them. The sensor outputs are subsequently used by the Automated Driving System (ADS) onboard the vehicle to make decisions that affect its trajectory and how it interacts with the physical world. The main sensor technologies being utilized for sensing and perception (S\&P) are LiDAR (Light Detection and Ranging), camera,  RADAR (Radio Detection and Ranging), and ultrasound. Different environmental parameters would have different effects on the performance of each sensor, thereby affecting the S\&P and decision-making (DM) of an AV.

In this publication, we explore the effects of different environmental parameters on LiDARs and cameras, leading us to conduct a study to better understand the impact of several of these parameters on LiDAR performance. From the experiments undertaken, the goal is to identify some of the weaknesses and challenges that a LiDAR may face when an AV is using it. This informs AV regulators in Singapore of the effects of different environmental parameters on AV sensors so that they can determine testing standards and specifications which will assess the  adequacy of LiDAR systems installed for local AV operations more robustly. Our approach adopts the LiDAR test methodology first developed in the Urban Mobility Grand Challenge (UMGC-L010) \textit{White Paper on LiDAR performance against selected Automotive Paints}. The intensity values of the entire point cloud are then obtained as output from the LiDAR, which enables us to evaluate the average reflected intensity and variance of the reflected intensity of the test objects. 

The tests were conducted with automotive paint panels as the test objects and a single Velodyne VLS-128 LiDAR sensor. The environmental parameters we have chosen as the focus of our study and to vary are: types of automotive paint, the angle of the surface, the distance of the object, and both dry and wet surface conditions. The basic test setup was the same as the one described in our previous white paper \cite{umgc_whitepaper1}.

\section{Introduction}
\label{sec:introduction}
In recent years, the development of robotic vehicles equipped with a dedicated Automated Driving System (ADS) \cite{standard2018j3016}, commonly known as autonomous vehicles (AVs), has received significant attention from both the academic research community and mobility industry across the world. Automated Driving Systems on AVs perform the task of monitoring the driving environment to perform the dynamic driving task. The onboard sensing and perception (S\&P) module of an AV is therefore a crucial component that must function reliably and with acceptable quality and range of detection in order to ensure system safety. AVs typically rely on sensors such as LiDAR (Light Detection and Ranging), visible light cameras, RADAR (Radio Detection and Ranging), and ultrasound to sense and perceive the world around them. The data from these sensors is typically fused together to generate a comprehensive \textit{object-list} of static and dynamic obstacles around the vehicle. The sensor data fusion ensures that the ADS has good spatial awareness with all-around coverage.

Numerous studies have detailed the behavior of camera systems under varied weather and environmental conditions and their response toward different materials, surfaces, and textures. Moreover, most camera systems used for AV applications function in the visible light spectrum and have a similar range of color vision as the human eye. Automotive LiDARs, on the other hand, typically operate in the infrared (IR) spectrum. There have been comparatively fewer studies carried out to investigate the performance of LiDAR systems in different weather and environmental conditions and also against automotive paint or other such materials, surfaces and textures.

Like humans `see' the world around them to make driving decisions, AVs use their sensors to `see' obstacles and vehicles around them. When AV sensors face different environmental conditions, it may be challenging for the sensors to 'see' obstacles and vehicles around them due to reduced visibility. Therefore, the degradation in visibility will have a detrimental effect on AV performance, particularly for the  S\&P and decision-making (DM) modules. This white paper will explore automotive cameras and LiDAR sensors; and the impact of various environmental parameters on these sensors.

\subsection{Automotive LiDARs}
The sensors for automated driving can be split into two groups: active sensors and passive sensors. LiDAR and radar sensors are classified as active sensors as they emit the signals they intend to receive back as reflections, whereas cameras are generally passive sensors, receiving signals originating from independent sources ie. natural or artificial light reflected from objects \cite{Beiker2018}. LiDARs stand out because it is integral to most AV perception systems and compensates for the weaknesses of camera and radar sensors. It complements camera-based perception algorithms by providing detailed range measurements, which may be limited otherwise, even with stereo cameras, and may not be possible with a single camera. Many commercially available and open-source Automated Driving (AD) stacks, such as Baidu Apollo \cite{Apollo_perception} or Autoware \cite{raju2019_autoware_apollo}, use LiDAR sensor data to generate a real-time object map of other road users and obstacles around the AV.

\subsubsection{Working principle}
LiDAR sensors \cite{li2020lidar} illuminate the surroundings by emitting laser light beams from an emitter, typically in the infrared (IR) spectrum of 903-1550nm. This region of the electromagnetic spectrum is not visible to the human eye \cite{li2020lidar}. The LiDAR sensor then uses a detector to detect reflected laser beams, generating a point cloud consisting of time-of-flight (ToF) calculated distances and reflected intensities of all objects around it \cite{behroozpour2017lidar}. A point cloud is a set of points in 3D space which represents the real world as seen by the LiDAR measurements.

\begin{figure}[htb]
    \centering
    \centerline{
        \includegraphics[scale=0.50]{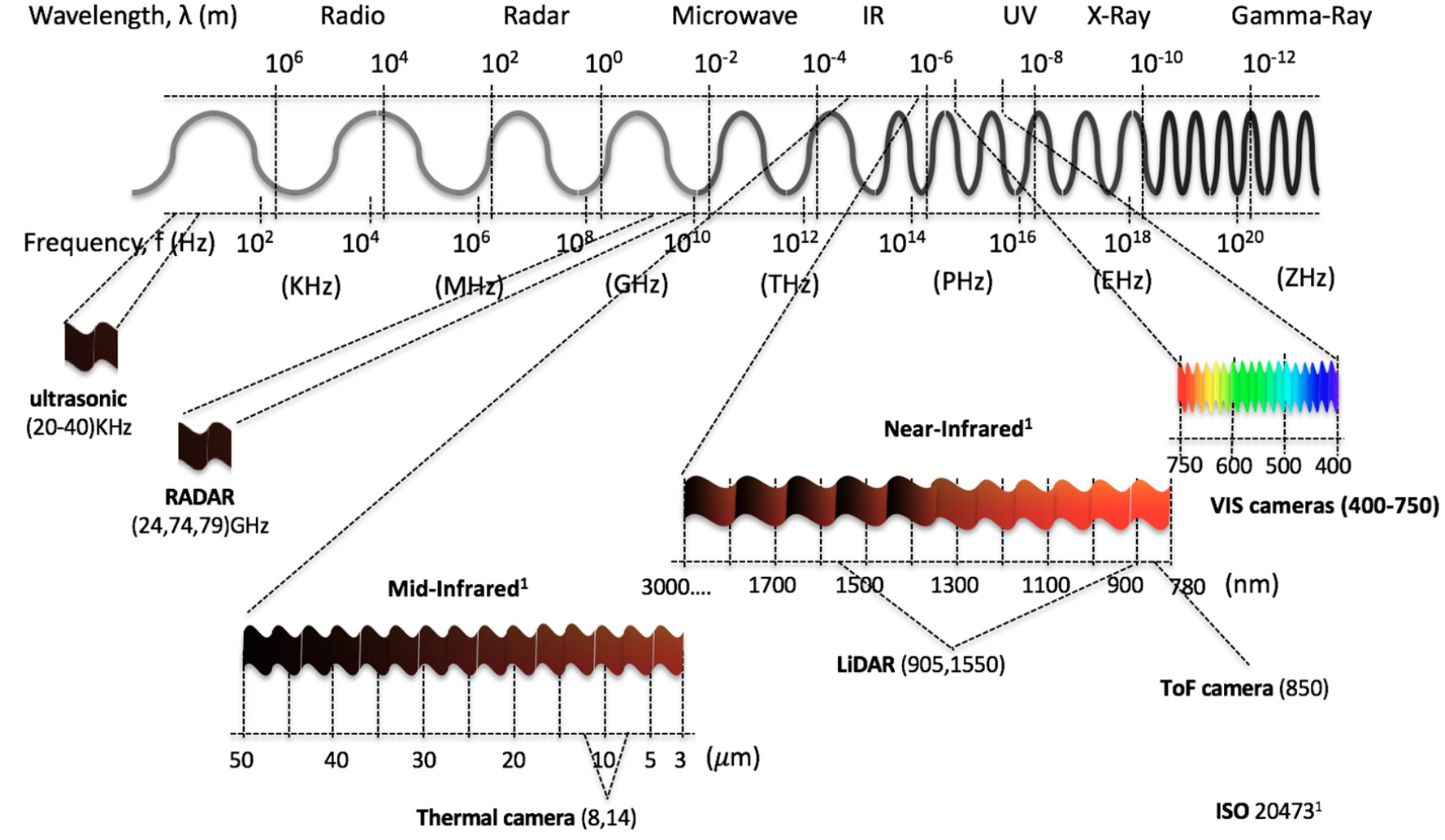}}
    \caption{Spectrum of electromagnetic radiation (Reproduced from \cite{rosique2019systematic})}
    \label{fig:spectrum}
\end{figure}

There are two main methods of conducting ToF measurements to calculate object distances. This can be done by measuring the actual time-of-flight of the laser beam based on the time delay between the pulse leaving the emitter and being received by the detector. Another way is to measure the phase shift between the emitted and received signals.

\begin{figure}[htb]
\captionsetup{justification=centering}
    \centering
    \includegraphics[scale=0.5]{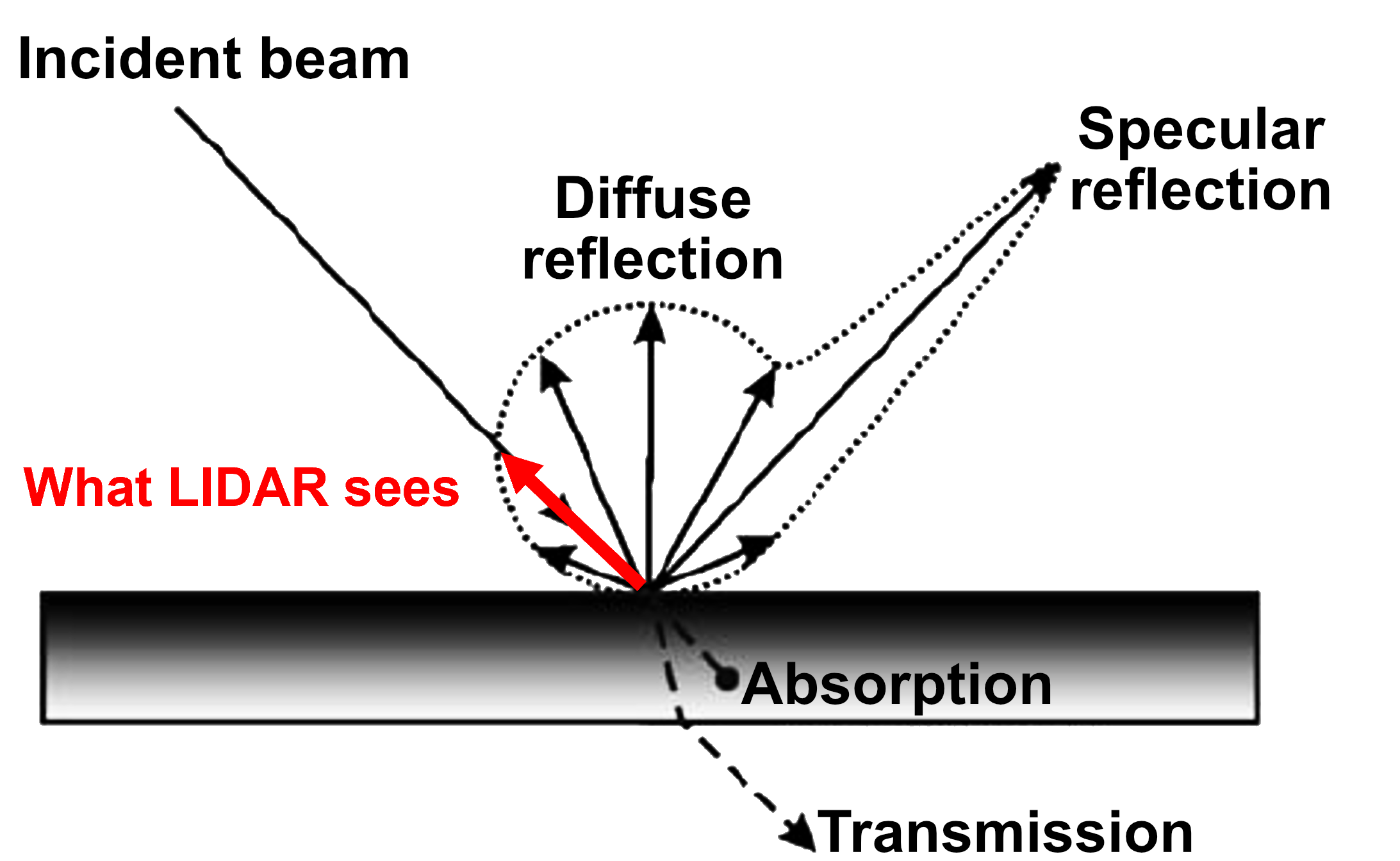}
    \caption{The LiDAR sensor primarily detects the incident reflection.\\At close distances, the diffuse and specular reflections may also become visible.}
    \label{fig:LiDAR_reflection}
\end{figure}

Optically, LiDAR works on the principle of diffuse reflection, whereby a ray of light is reflected back at the same angle as the incident beam. This phenomenon is illustrated in Fig.~\ref{fig:LiDAR_reflection}. The LiDAR measures the ToF of a pulse of light from the moment it gets emitted from a laser diode until it is received by a detector after it is reflected by a reflective surface. The LiDAR detectors may also pick up beams from the specular reflection at very close distances. The range of surrounding objects can then be accurately calculated from these ToF measurements and a point-cloud of the environment is constructed.

\subsubsection{Data outputs}
The data output from LiDARs varies based on manufacturer specifications. Velodyne LiDARs typically return a value known as \textit{calibrated reflectivity}, which is described as ``reflectivity values returned based on NIST\cite{NIST}-calibrated reflectivity targets'' from their factory \cite{vlp16_usermanual}. This is nothing but the returned power value of the reflected light as measured against the calibrated reflectivity NIST test targets. Other LiDAR OEMs, such as Robosense and Quanergy also return the reflected intensity or reflectivity value along with the X, Y and Z coordinates of each point in space. Sensor manufacturers also implement comprehensive and proprietary post-processing algorithms within the sensors to ensure optimal performance under different lighting conditions.

\subsubsection{Types of LiDAR, advantages and disadvantages}
There are various types of LiDARs available in the market today. Some of the types of LiDARs used for automated driving and Advanced Driver Assistance Systems (ADAS) applications are:
\begin{itemize}
    \item Electromechanical LiDAR
    \item Solid state LiDAR
\end{itemize}

Within each of these categories, there exists both 2D and 3D LiDARs. The following paragraphs will explain different LiDAR technologies, their advantages and their disadvantages.

\textbf{Electromechanical LiDAR -} 
Electromechanical LiDARs collect data over a wide area of up to 360 degrees by physically rotating a laser and receiver assembly, or by using a rotating mirror to steer a light beam. Electromechanical LiDARs use powerful, collimated lasers that concentrate the return signal on the detector through highly focused optics \cite{whylidar}.

Electromechanical LiDARs can be designed to return either a 2D or 3D point cloud. In a 2D LiDAR, there may only be a single spinning laser beam or a linear array of laser beams arranged in a single axis. The resulting point cloud may have up to a 360~\degree coverage, but may only provide information about obstacles in a single plane. 2D LiDARs usually provide a low resolution segment-based detection map. This makes them useful for near field applications to detect obstacles within the immediate vicinity of the AV. It is suitable for performing simple detection and ranging tasks on planar surfaces and are not suitable for detailed object detection and all-round perception for any autonomous system \cite{Infographic}.

In a 3D electromechanical LiDAR, the emitter-detector array may be arranged either in a vertical linear fashion, or as a planar array around the rotational axis. High resolution sensors such as the Velodyne VLS-128 have a complex laser emitter-detector pattern in order to accommodate the large number of emitter-detector pairs and to improve scanning resolution. 

\textbf{Solid State LiDAR -}
Solid State LiDARs are designed and built without motorized mechanical scanning. It has no moving mechanical parts and works by providing complete, instantaneous scene illumination through emitted laser flashes and capture incremental insights on objects \cite{whylidar}.

2D solid state LiDARs may have only a linear array of emitter-detector pairs. With advances in MEMS technologies and improved manufacturing techniques, solid state 3D LiDARs are also being developed. These typically rely on a 2-dimensional array of emitter-detectors which emit laser flashes in order to scan a 3D area in front of the sensor and generate the point cloud.

In general, 3D LiDARs are more suitable for detailed analysis of an environment and thus are suited for tasks such as object detection and collision avoidance on automated vehicles \cite{Infographic}. Since they are used for detailed perception tasks, 3D LiDARs typically are designed with the objective of longer detection ranges achieved using higher laser emitter power and detector amplification.

\subsection{Automotive cameras}
Camera sensors are generally passive sensors that receive signals that originate from independent sources ie. natural or artificial light reflected from objects \cite{Beiker2018}. Automotive cameras are cameras used by a vehicle as part of its Advanced Driver Assistance System (ADAS) or for its highly automated driving functions. It plays an important role in the majority of AV perception systems. Cameras provide images as its output, and after image processing together with advance computer vision algorithms, is able to perform object detection and classification effectively. Many commercially available and open-source Automated Driving (AD) stacks such as Baidu Apollo \cite{Apollo_perception} or Autoware \cite{raju2019_autoware_apollo}, use automotive camera sensor data to carry out tasks such as the detection of static, dynamic obstacles and recognition of traffic light states.

\subsubsection{Working principle}
Camera sensors for automotive use cases usually work in the visible light range (wavelength 400nm - 780nm) of the electromagnetic spectrum as seen in Fig. \ref{fig:spectrum}. These cameras are classified as visible light cameras. Environmental lighting within these range of wavelengths, whether it is natural or artificial lighting that originate from independent sources, will strike an object of interest. A camera will then receive light reflected off this object and capture it when it hits an imaging sensor, after passing through a lens. This process is illustrated in Fig. \ref{fig:camera_wp} below.
\begin{figure}[htb]
\captionsetup{justification=centering}
    \centering
    \includegraphics[scale=0.65]{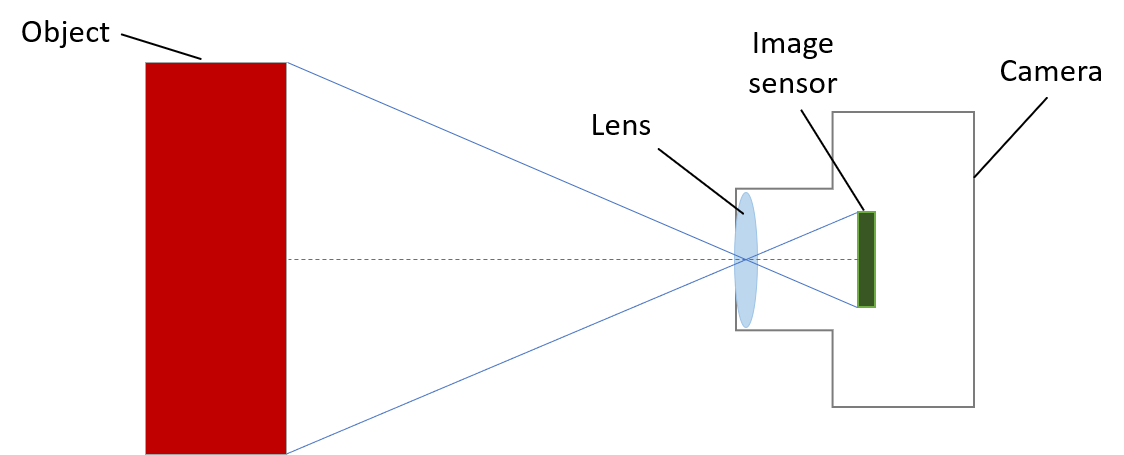}
    \caption{Working principle of a digital camera}
    \label{fig:camera_wp}
\end{figure}

Light captured by the image sensor will be subsequently divided into three bands or channels: Red (R), Green (G), Blue (B), which will be coded separately. The result is a digital RGB image output. Taking multiple image frames and combining them over time will give a video stream which is used by an AV's perception system to obtain surrounding information.

Utilising two cameras with known focal distances also allows for stereoscopic vision to be carried out by adding depth (D) information to an image. The additional fourth channel will give the image RGBD data. This will supply a 3-dimensional representation of the scene around the vehicle to an AV's perception system \cite{rosique2019systematic}. 

Less commonly seen and used on AVs but more widely used in other industries is another type of camera known as the infrared (IR) camera. These cameras work in the IR range (wavelength 780nm - 1mm) of the electromagnetic spectrum. They follow the same working principle as visible light cameras, but instead of capturing light from the visible light range, it captures light from the IR range when it bounces off an object. 

\subsubsection{Data outputs}
The data outputs for an automotive visible light camera are normally digital RGB images, converted from internal raw images of the same  or another colour space specific to the image sensor of the camera. The RGB images have a fixed number of pixels based on the desired image size or resolution set during sensor configuration. Each pixel will have separate Red, Green, Blue channel information, giving it and the entire image colour. This image will be in the standard RGB colour space, similar to how the human eyes are able to visually perceive the surrounding world. Camera sensor manufacturers can also implement image post-processing algorithms within the sensors themselves to provide other relevant data such as image semantic segmentation information or depth information if multiple cameras are used.  

For IR cameras, the data outputs are different as compared to visible light camera outputs. IR cameras produce grey scale images. This is due to the image sensor being used that generates monochrome raw images when capturing light in the IR range.

\subsubsection{Types of Cameras, advantages and disadvantages}
Camera sensors just like other sensor technologies being utilised on AVs have its advantages and disadvantages. There are various types of cameras available in the market today. Some of the types of cameras used for automated driving and Advanced Driver Assistance Systems (ADAS) applications are:

\begin{itemize}
    \item Visible light camera
    \item Infrared (IR) camera
\end{itemize}

The following paragraphs will explain different camera technologies, its advantages and disadvantages.

\textbf{Visible light camera -}
Visible light cameras as mentioned previously, works in the visible light range by collecting visible light on an image sensor and processing it to produce an RGB image. The image sensor has traditionally been implemented with one of two technologies available, either Charge-coupled Device (CCD) or Complementary Metal Oxide Semiconductor (CMOS) \cite{rosique2019systematic}.

Such a camera is generally utilised for tasks like lane detection, obstacle detection and classification by an AD stack. This is because cameras are good for classification of objects when used in conjunction with image processing algorithms. The camera is also able to provide additional information about the environment through its colour and high resolution images.

However, visible light cameras do have several disadvantages. They generate a huge amount of data output through the constant stream of images, which makes it computationally expensive for the processing system of an ADS. Visible light cameras are also known to be sensitive to environment lighting conditions, performing worse at night and in other dark surrounding environments. The cameras are highly susceptible to poor performance in bad weather conditions too \cite{van2018autonomous}.

\textbf{Infrared (IR) camera -}
Infrared cameras as discussed earlier, works in the infrared light range by capturing infrared light on an image sensor and processing it to produce a grey scale image. These cameras are used to complement visible light cameras and not meant to replace them fully. They are able to perform better in low light conditions as compared to visible light cameras. Nonetheless, infrared cameras do not provide colour information of the environment and its output images are typically of lower resolution.

\section{Environmental parameters for AV sensors}\label{sec:parameters}
\subsection{Lighting conditions}
Lighting conditions refer to the amount of light in the environment that an AV operates in.  This can vary from low lighting levels at night  to high lighting levels during the day on open roads. The lighting conditions of the environment around an AV is known to have an effect on some of the sensors used for AV sensing and perception. In this subsection, a closer look is taken with regards to how lighting conditions affect LiDAR and camera sensors and their outputs.

\subsubsection{Effects of lighting conditions on LiDAR}
LiDARs are active sensors that emit laser light beams to measure and detect objects. Since, LiDARs generate its own light source of specific wavelengths, the effects of environmental lighting on LiDARs and its output data can be considered minimal. In a study done by \cite{anand2022evaluation} on the quality of LiDAR data in varying ambient light, a Velodyne VLP-32C LiDAR was tested outdoors during different times of the day to cover for varying lighting conditions. Four time periods were chosen to conduct the test. These were early morning and evening to represent the transition time when light level is diminishing, mid-day for maximum brightness and night for no light conditions. The LiDAR was tested against a person acting as an object at distances between 5m to 30m. The LiDAR parameters recorded are the number of points and the point density of the object. From the experimental results, the number of object points decreases as the distance of the object from the LiDAR increases and this trend is similar across all time periods. Likewise, the point density decreases with increasing object distance, a trend also similar across the four time periods. Hence, these results establishes that a LiDAR is invariant to changes in external light and it can be concluded that the effect of ambient light on LiDAR data is also negligible.

\subsubsection{Effects of lighting conditions on Camera}
Cameras are passive sensors that depend on surrounding light originating from independent sources in order for it to sense and detect objects of interest. As such, a sufficient amount of environmental light is crucial to the performance of any cameras used for AV applications. There would be a noticeable degradation of camera output data if light levels are too low or too high. When lighting levels are too low, the number of light rays hitting an image sensor are lesser. This results in images produced by the camera to be darker and noisier, and might have the effect of distorting the details in an image. On the other hand, if lighting levels are too high, images produced by the camera will turn out to be too bright and some details in an image might be lost. Therefore, lighting conditions have a significant impact on camera performance and data output.

\subsection{Automotive paint}
Automotive paints specifically refer to the paints used by the automotive industry and applied onto vehicles for the purposes of protecting external body panels and improving the appearance of the vehicle \cite{toda2012automotive}. Automotive paints are usually applied in multiple layers onto the panel of a vehicle body during the production process. It consists of the substrate, which is the panel itself, made of materials such as steel, aluminium, plastics or more recently, composites. A primer is applied to the panel which bonds to it to produce a uniform coating thickness. Next, a base coat, also known as a colour coat is applied onto the panel, giving the surface its colour. Finally, a top coat, also known as a clear coat is applied on top of the base coat to act as a protective coating, mainly to protect the base coat from ultraviolet light damage \cite{toda2012automotive}. 

\subsubsection{Types of automotive paints}
Automotive paints generally fall into categories such as metallic, non-metallic, glossy and matte. The difference between metallic and non-metallic paints lie in the addition of metallic pigments for metallic paints. Metallic pigments consist of very thin platelet-shaped particles made out of aluminum or bronze and they are added on top of the base coat, before applying the top coat. The difference between glossy and matte paints lies in the type of top coat being applied. The top coat can either have a glossy or matte finish. This gives rise to varying abilities of a paint surface to reflect light \cite{automotive_handbk}.

\subsubsection{Effects of paints on LiDAR}
The paint colour and reflectivity of a vehicle body surface usually has an impact on LiDARs. This is largely due to the ability of the target material to reflect laser light back to the detectors in the LiDAR sensor. In research done to characterise LiDAR sensors for 3D perception, it was found that the reflected intensity of black coloured surfaces is typically lower when compared to a white one \cite{wang2018characterization}.

Additionally, material reflectivity also has an impact on LiDARs. Metallic surfaces or material coated with metallic paint presents challenges to LiDAR sensors and thus affects LiDAR based perception. In a study \cite{pomerleau2012noise} to verify the impact of reflectivity, it was established that reflective surfaces such as an aluminium plate poses three challenges to LiDAR sensors. First, when the incident angle is large, most of the energy is not reflected back to the sensor, which can lead to missing measurements. Second, there is also a probability that the laser beam gets reflected to another surface leading to an overestimation of depth. Finally, reflective plates exhibit a larger spectrum of reflected intensity, which seems to create systematic error producing wave patterns. All these sensing challenges could affect LiDAR based AV perception systems, preventing them from carrying out the goal of performing accurate object detection.

\subsubsection{Effects of paints on Camera}
Automotive paints are not known to have an effect on cameras under normal lighting conditions. A camera will be able to detect paint panels of a vehicle the same way regardless of its colour. However, under poor or low light conditions, darker paint panels are harder to detect by the camera as compared to lighter coloured paint panels. This is because darker paint panels tend to absorb more visible light and reflect lesser light to the camera, providing a smaller contrast to the dark environment.

\subsection{Angle of an object's surface}
The angle of an object's surface predominantly refers to the orientation of an object in the surrounding environment of an AV. The way that an object is orientated, might mean that an AV's sensor is facing it perpendicularly or in parallel to it at the other extreme. In this subsection, an analysis of how different aspects of an object can have varying effects on both LiDAR and camera sensor outputs.

\subsubsection{Effects of surface angles on LiDAR}
\begin{figure}[htb]
\captionsetup{justification=centering}
    \centering
    \includegraphics[scale=0.4]{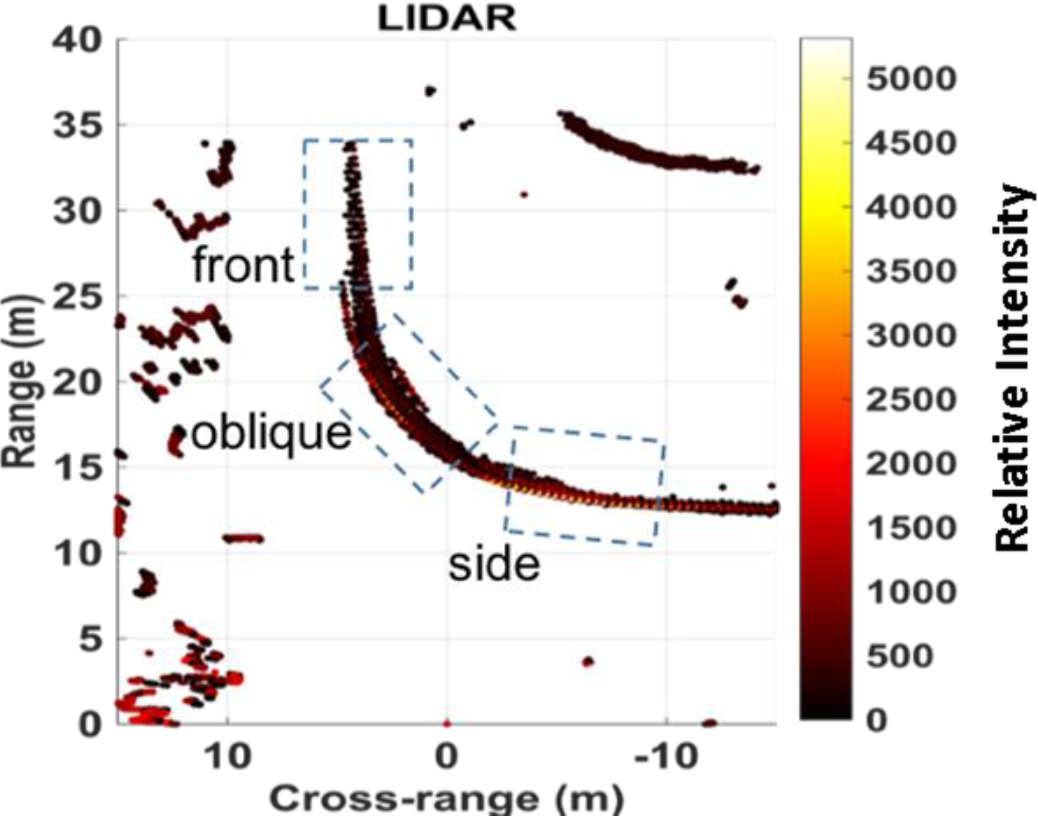}
    \caption{LiDAR relative reflected intensity data collected in the study \cite{xique2018evaluating}}
    \label{fig:LiDAR_aspect_graph}
\end{figure}

LiDAR sensors as mentioned in previous sections are an active sensor. This means that a LiDAR emits laser light beams and measures the power of the light reflected off the surface of an object. When the orientation of the object changes, its surface angles will also change relative to the LiDAR. This affects the amount and power of light being reflected back to the LiDAR. A research was conducted by \cite{xique2018evaluating} to evaluate complementary strengths and weaknesses of ADAS sensors. In their experimental setup, a scenario whereby an oncoming vehicle carries out a left turn while a stationary ego equipped with sensors observes the oncoming vehicle was implemented. The scenario enabled the researchers to gather data of the oncoming vehicle over several viewing aspects such as the front, oblique and side aspects. These data is shown in Fig. \ref{fig:LiDAR_aspect_graph}.

From the data, LiDAR results were computed and were found that at side and oblique vehicle orientations, when majority of its body panels are either perpendicular or slightly off perpendicular to the sensor, reflected intensity was higher and almost the same. However, when the test vehicle is at frontal orientation, when most of its body panels are parallel to the LiDAR, performance of the sensor, measured as the LiDAR's relative intensity, decreases by about $70\%$. Hence, it is certain that a target object's surface angles has an effect on LiDAR data outputs.

\subsubsection{Effects of surface angles on Camera}
The orientation of an object and its surface angles, are not known to have an effect on cameras and its ability to detect the object. However, there would be lesser information of the object in the resulting output images. This might cause AV perception systems to wrongly classify an object if fed with such images as input. 

\subsection{Distance of object}
Distance of objects refer to the distance between static/dynamic objects in the environment and the AV sensors themselves. The distance of different objects to an AV is known to have an effect on the majority of the sensors used for AV sensing and perception. In this subsection, a review is done on how distance of objects affect LiDAR and camera sensors and their outputs.

\subsubsection{Effects of object distances on LiDAR}
Object distances does have an effect on LiDAR performance based on data gathered from past research. This subsection will mainly cover the impact of object distances on 3D LiDARs which is the type of LiDAR most widely used on AVs. For 3D LiDARs, aside from horizontal, vertical and range information, the LiDARs also have an intensity channel which is established by calculating the return power of the laser and thus a material's reflectivity. Most LiDAR's intensity will be calibrated by distance to counteract the natural decrease in returned power, depending on hardware and on-board processing \cite{Lambert2020}. However, what is of greater concern, especially for lower vertical resolution LiDARs, is that at further distances, LiDAR measurements become more sparse and point cloud output is less dense. At a longer distance, the LiDAR beam layers diverges greatly. The gap between the layers increases and the number of points of the point cloud that fall onto the target object decreases \cite{marti2019review}. This means that at further distances, LiDAR output data will suffer from a diminishing amount of information of the surrounding objects.     

\subsubsection{Effects of object distances on Camera} 
Varying object distances have differing effects on a camera and its output. The optimal distance for an object in front of a camera is dependent on the focal length of the camera lens. If an object is too near, the object may appear blurry in the output image. If an object is too far, the size of the object becomes very small in the output image and some object details will be lost. This might lead to a wrong classification of the object when the image is fed to the AV perception module.

\subsection{Surface conditions (Dry/Wet)}
As AV sensors are used to detect obstacles in an outdoor environment, they may encounter objects that have a dry or wet surface, due to weather conditions. The bulk of time, AVs will be faced with obstacles having dry surfaces. Occasionally, after rain or during times with slightly colder temperatures when water vapour from the atmosphere condenses onto an object surfaces, an AV will face obstacles with wet surfaces. In this subsection, the effects of both dry and wet surface conditions on AV sensors are investigated.

\subsubsection{Effects of surface conditions on LiDAR}
Surface conditions of an object have a large influence on LiDAR performance and output. This is specifically because of its impact on target object surface reflectivity. When an object surface is dry, the factors affecting surface reflectivity is mainly the material and the colour of the object surface. However, when an object surface is wet, the water droplets resting on top of it plays a significant role in altering the object's surface reflectivity properties. The water droplets might cause a scattering of the laser beams, so much as to result in more diffuse reflections that the LiDAR cannot see like in Fig. \ref{fig:LiDAR_reflection}. Therefore, the reflected intensity of the LiDAR point cloud output will decrease.

\subsubsection{Effects of surface conditions on Camera}
The surface conditions of an object have a limited effect on camera performance and output. In terms of object detection, a camera will still be able to "see" an object regardless of whether its surface is dry or wet. A minor effect that can be observed when a wet surface is under strong sunlight is that the object in the output image might appear to be shiny, because of the way the water droplets reflect light to the camera. Nonetheless, this has little effect on camera performance.

\clearpage

\section{LiDAR test methodology with various environmental parameters}
In literature, several studies and experiments \cite{wang2018characterization}, \cite{pomerleau2012noise}, \cite{rasshofer2011influences}, \cite{carrea2016correction}, \cite{reflectivityofpaint} have focused on determining LiDAR performance under varying ambient conditions and for simple coloured materials across a variety of industry applications. This section will focus on the testing of automotive grade LiDAR against specific environmental parameters that an AV is likely to encounter, as mentioned in Chapter \ref{sec:parameters}. The section will cover the LiDAR testing methodology utilised and the test plan and inputs required to enable testing of LiDAR against different environmental parameters.

\subsection{LiDAR testing setup}
The LiDAR test setup employed for our testing of LiDAR against different environmental parameters has been largely adopted from the work done in the study documented in \textit{White Paper on LiDAR performance against selected Automotive Paints} \cite{umgc_whitepaper1}. In this subsection, we will briefly go through the testing methodology, and for a more comprehensive overview of the methodology, readers can refer to the white paper published from that study.

\begin{figure}[htb]
    \centering
    \centerline{
        \includegraphics[scale=0.45]{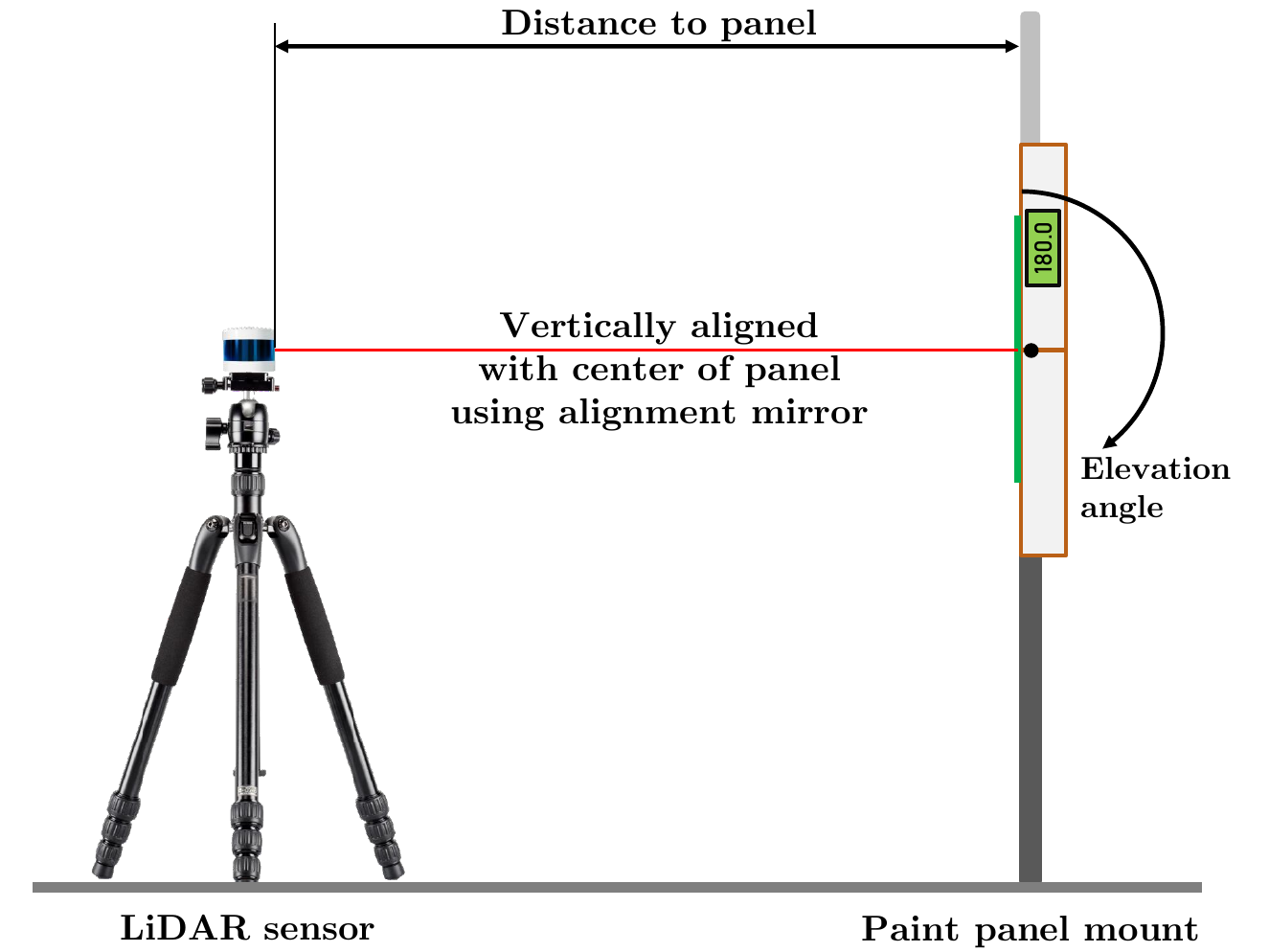}
    }
    \caption{Adopted LiDAR testing setup}
    \label{fig:test_setup}
\end{figure}

The test setup consists of the following as illustrated in Fig.~\ref{fig:test_setup}:
\begin{itemize}
    \item \textbf{Sensor under test} - The LiDAR sensor is mounted  on a height adjustable tripod. The height of the LiDAR sensor matches that of the panel mount and its alignment mirror as described subsequently.
    \item \textbf{Test targets} - Sample paint panels were obtained through a partnership with paint and coating supplier NIPSEA Technologies.  These sample panels are flat square nickel coated mild steel plates and are spray-painted manually by the supplier.
    \begin{itemize}
        \item Dimensions for a single panel are 50cm x 50cm due to the physical constraints of the manual spray booth.
        \item To test larger targets, 4 pieces can be combined to obtain a 1m x 1m test target.
    \end{itemize}
     \item \textbf{Panel mount} - The panel is mounted with a pivoting setup with adjustable elevation angles. The pivoting setup is developed from standard white board hardware with specific modifications made to improve test reliability and repeatability:

    \begin{itemize}
        \item The elevation angle can be measured using a digital angle gauge which is physically and securely attached to the pivot points of the test rig.
        \item The paint panel test targets are held in place on the white board surface using carefully positioned and aligned high-strength neodymium magnets. These magnets help to align the panels and also hold them flat against the surface ensuring that the panels are perfectly flat.
        \item The white board surface also includes a mirror located at the centre of the pivot point. This mirror helps to align the sensor with the white board surface.
    \end{itemize}
    \item \textbf{LiDAR data capture computer} - A computer is connected to the LiDAR to capture in real-time the output point-cloud in order to analyze sensor performance against various test targets. The data capture computer runs ROS with custom-developed packages in order to automate measurement.
\end{itemize}
\clearpage

\subsection{Test plan and inputs}
\begin{figure}[htb]
    \centering
    \centerline{
        \includegraphics[scale=0.5]{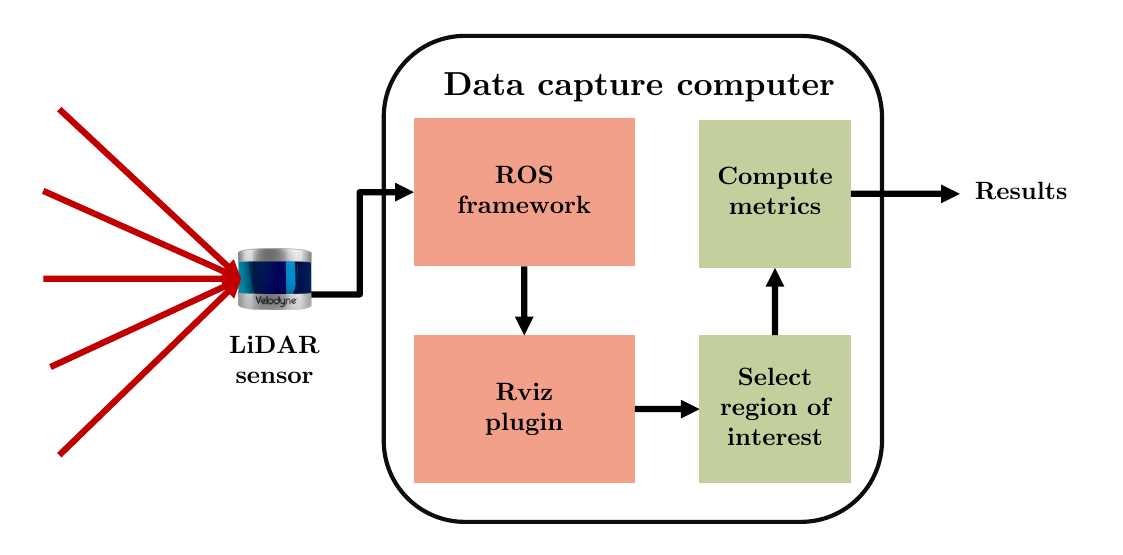}
    }
    \caption{Overview of the data capture software framework}
    \label{fig:Data_capture}
\end{figure}

A LiDAR sensor will be tested in controlled environments against common automotive paints and coatings applied on a test target, while varying properties such as lighting, distance, orientation angle and surface conditions of the test target. In order to keep external variables (such as other environmental effects and stray light) to a minimum and ensure controllability and repeatability, the experiments will be conducted under the following environmental conditions and with the listed equipment:

\begin{itemize}
    \item \textbf{Location} 
        \begin{itemize}
            \item Indoors (In a large indoor laboratory): 40m x 11m x 7m
            \item Outdoors (Mainly for verification testing): 25m x 5.5m
        \end{itemize}
    \item \textbf{Lighting}
        \begin{itemize}
            \item Office fluorescent lighting (Indoors)
            \item Sunlight (Outdoors)
        \end{itemize}
    \item \textbf{Target surface conditions}
        \begin{itemize}
            \item Dry
            \item Wet (Manual spraying of water droplets onto surface)
        \end{itemize}

    \item \textbf{Sensor under test} - It is important to test an automotive grade LiDAR sensor in order to gain a thorough understanding of the challenges faced by AV sensing and perception systems against other vehicles, obstacles and environmental parameters on the road. The LiDAR that would be part of the experimental setup is as follows:
    
        \textbf{Electromechanical LiDAR} 
        \begin{itemize}
                \item Velodyne VLS-128 (128 channel, long range) - in `Strongest return' mode:
            \begin{itemize}
                \item Horizontal FoV: 360\degree
                \item Vertical FoV: -25\degree ~to +15\degree, non-linear beam pattern
                \item Rotation speed: 540 rpm
                \item Mount position: Horizontally mounted
            \end{itemize}
        \end{itemize}
    \item \textbf{Test targets} - Automotive grade paint panels were used as test targets. These are metal panels painted with automotive paint and affixed to the panel mount while being held in place by its magnets. The target specifications are listed and described in Table \ref{table:NIPSEA_paints}. 

The sample paint panel test targets also needs to be tested with different parameters and
physical states as highlighted below:

\begin{enumerate}
    \item \textbf{Varying elevation angle} - From 0\degree~to 60\degree~at 15\degree~intervals
        \begin{itemize}     
        \item 0\degree~being normal to the LiDAR beams  
        \end{itemize}
    
    \item \textbf{Varying target distance}
    \begin{itemize}     
       \item 5.0m, 10.0m, 20.0m, 30.0m
    \end{itemize}
\end{enumerate}

\begin{table}[htb]
\centering
\begin{tabular}{|l|c|c|l|l|}
\hline
No. & Colour                                                                       & \multicolumn{1}{l|}{Surface Finish} & Panel Code  & Remarks              \\ \hline
1   & \multirow{5}{*}{Black}                                                       & Gloss                               & SB-Gloss*    &                      \\ \cline{1-1} \cline{3-5} 
2   &                                                                              & Gloss                               & FB1-Gloss**   & LiDAR functionalised \\ \cline{1-1} \cline{3-5} 
3   &                                                                              & Matte                               & SB-Matt*     &                      \\ \cline{1-1} \cline{3-5} 
4   &                                                                              & Matte                               & FB1-Matt**    & LiDAR functionalised \\ \cline{1-1} \cline{3-5} 
5   &                                                                              & Gloss                               & FB4-Gloss**   & LiDAR functionalised \\ \hline
6   & White                                                                        & Gloss                               & SW-Gloss    &                      \\ \hline
7   & \multirow{2}{*}{Blue}                                                        & Gloss                               & CDSBL-Gloss &                      \\ \cline{1-1} \cline{3-5} 
8   &                                                                              & Gloss                               & CDFBL-Gloss & LiDAR functionalised \\ \hline
9   & \multirow{2}{*}{\begin{tabular}[c]{@{}c@{}}Red\\ (Metallic)\end{tabular}}    & Gloss                               & TCSRM-Gloss &                      \\ \cline{1-1} \cline{3-5} 
10  &                                                                              & Gloss                               & TCFRM-Gloss & LiDAR functionalised \\ \hline
11  & Green                                                                        & Gloss                               & SMRTG-Gloss &                      \\ \hline
12  & \multirow{2}{*}{\begin{tabular}[c]{@{}c@{}}Silver\\ (Metallic)\end{tabular}} & Gloss                               & TSSM-Gloss  &                      \\ \cline{1-1} \cline{3-5} 
13  &                                                                              & Gloss                               & TFSM-Gloss  & LiDAR functionalised \\ \hline
\end{tabular}
\medskip
\medskip
\\
*SB / Standard black -- Black paint with standard black pigment\\
**FB / Functionalised black -- Variants of black paint with special LiDAR functionalised pigment
\caption{A summary of the tested paint panels from NIPSEA}
\label{table:NIPSEA_paints}
\end{table}

    \item\textbf{LiDAR data capture computer} - The test results were recorded using the LiDAR sensor data capture computer and its ROS-based framework (Fig.~\ref{fig:Data_capture}). The average intensity of the test panel region was then extracted from the point cloud. This ranges from 0 -- 255 for the Velodyne LiDAR.

\end{itemize}

\section{Results and analysis}

\subsection{Results}
The results consist of the experimental outcomes gathered from the tests done to find out the effects of various environmental parameters on the VLS-128 LiDAR sensor. The relevant panels for each test were mounted on the panel mount separately in order to obtain raw point cloud data and the reflected intensity. Three readings were taken for each type of panel and the average reflected intensity was also computed. This was done in a lab indoors at the selected distances and for multiple elevation angle orientations as mentioned previously. Whenever we wanted to replicate wet surface conditions, water was manually sprayed onto the paint panel test targets, leaving water droplets on its surface. The results from the tests were consolidated into average reflected intensity and variance of reflected intensity trends across the different environmental parameters for easier comparison. These are further elaborated in the subsections below.

\subsubsection{Verifying effects of lighting}
\begin{figure}[h!]
    \centering
    \centerline{
        \includegraphics[scale=0.52]{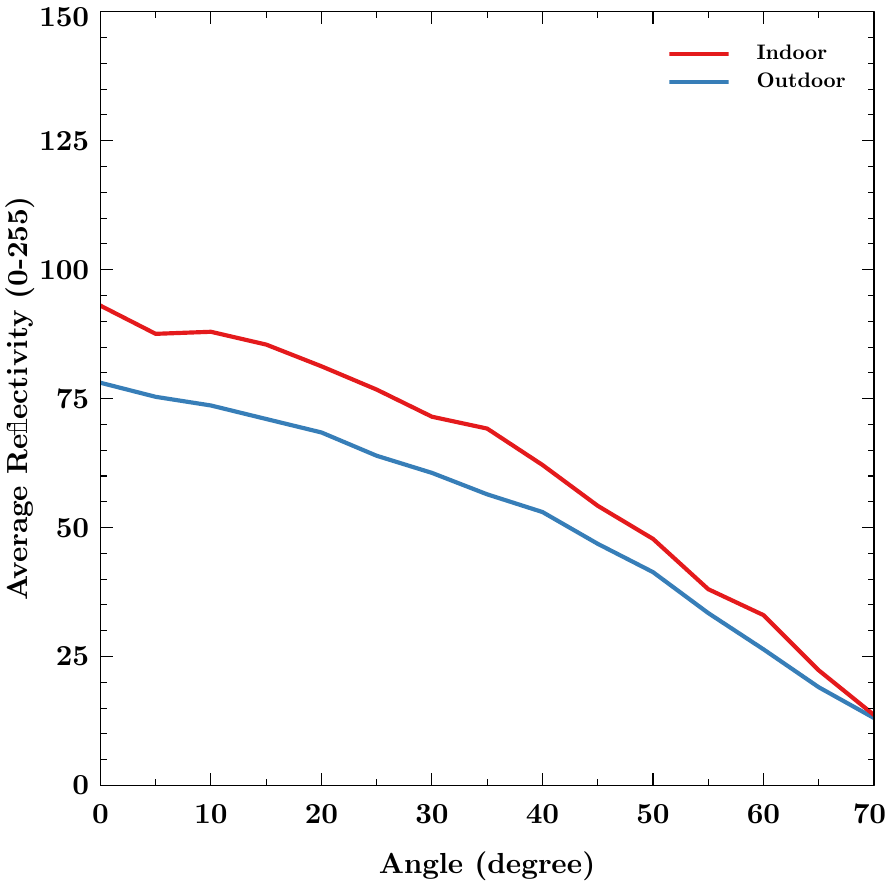}
    }
    \caption{VLS-128 Average reflectivity vs. Angle of incidence for SW\_Gloss at 10m different locations}
    \label{fig:SWGloss_outdoor}
\end{figure}

\begin{figure}[h!]
    \centering
	\captionsetup{justification=centering}
	\subfloat[]{\label{fig:SBGloss_outdoor} \includegraphics[width=0.48\columnwidth]{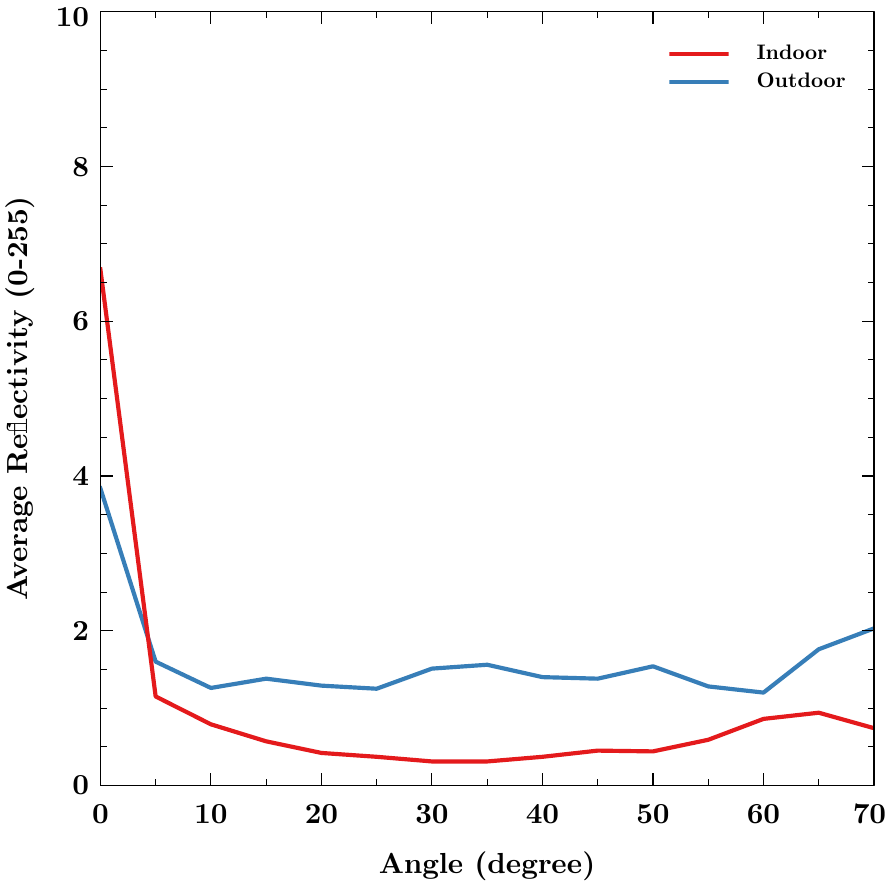}}
	\hspace{5pt}
	\subfloat[]{\label{fig:SBMatt_outdoor} \includegraphics[width=0.48\columnwidth]{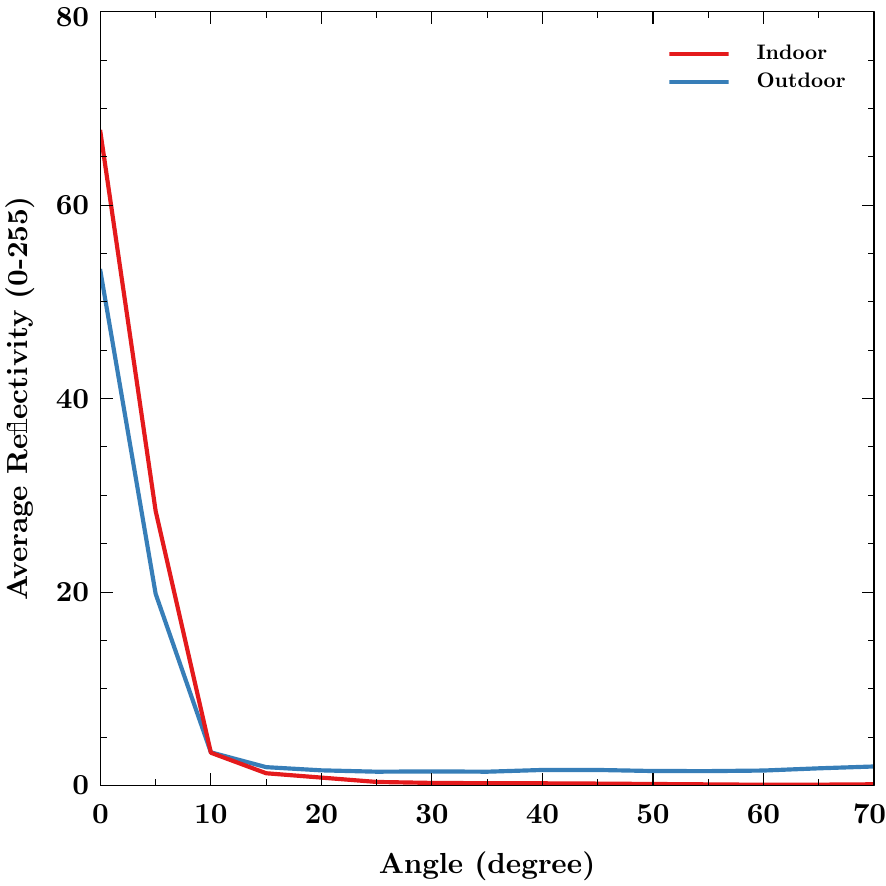}}
	\caption{VLS-128 Average reflectivity vs. Angle of incidence at 10m different locations \textbf{(a)} SB\_Gloss \textbf{(b)} SW\_Matt}
	\label{fig:SB_outdoor_compare}
\end{figure}

Based on our literature review of various environmental parameters for AV sensors done in Chapter \ref{sec:parameters}, it was established that ambient lighting conditions had negligible effects on LiDAR performance and output. To verify this, we carried out  testing of the VLS-128 LiDAR outdoors and around noon, when the level of sunlight is at its maximum. Three different panel types (SW-Gloss, SB-Gloss and SB-Matt) were placed at a distance of 10m from the LiDAR and their elevation angles were varied from 0\degree ~to 70\degree ~at 5\degree ~intervals. Subsequently, the average return intensities from the test panels were recorded. Thereafter, we compared it with the data collected from the same test panels taken during indoor testing, under fluorescent lighting. Fluorescent light has wavelength in the visible light range, which means it has no effects on a LiDAR. Hence, data from the indoor tests is representative of low light conditions a LiDAR might face when it is operating on an AV at night.  A comparison of indoor and outdoor data for the three different panels are shown in  \autoref{fig:SWGloss_outdoor} and \autoref{fig:SB_outdoor_compare}.

From the figures above and below, we can see that the indoor (no sunlight) and outdoor (max sunlight) trends are similar, with both showing a gradual decrease in intensity as angle increases for the three panel types. Therefore, the results confirm that ambient lighting and the amount of light in the environment has negligible effects on LiDAR performance.

\subsubsection{Tests against automotive paints}
The tests against automotive paints were conducted using all the paint types listed in Table \ref{table:NIPSEA_paints}, which consists of  LiDAR functionalised\footnote{LiDAR functionalised paints have a special functionalised base coat applied by NIPSEA, this adjusts the paint’s colour coating system and calibrating it to achieve different Near Infrared Radiation (NIR) reflectance values}, non-LiDAR functionalised, metallic, non-metallic, glossy and matte paints.

\begin{figure}[h!]
    \centering
	\captionsetup{justification=centering}
	\subfloat[]{\label{fig:ave_functional} \includegraphics[width=0.48\columnwidth]{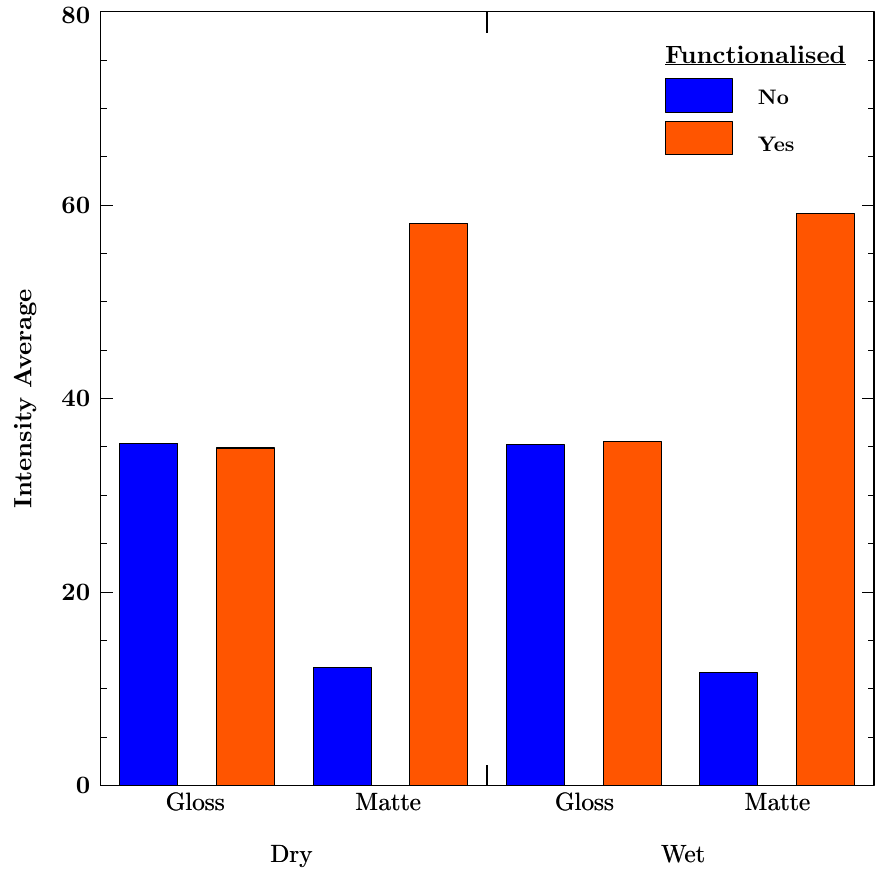}}
	\hspace{5pt}
	\subfloat[]{\label{fig:var_functional} \includegraphics[width=0.48\columnwidth]{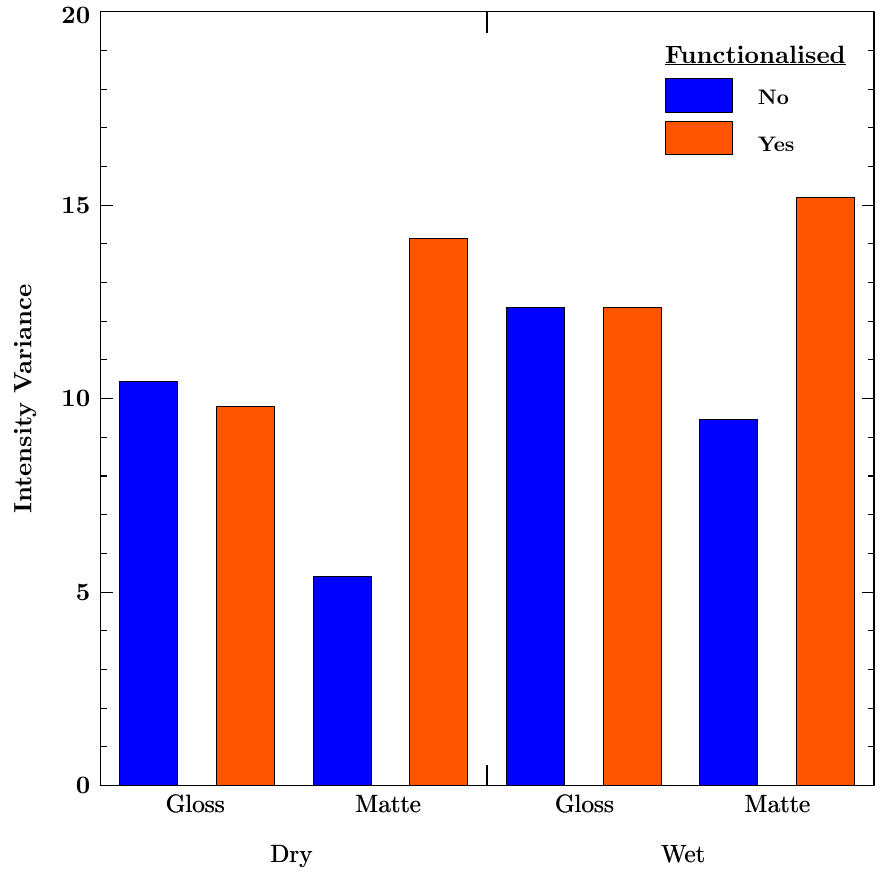}}
	\caption{LiDAR functionalised paint type comparison \textbf{(a)} Based on Intensity Average  \textbf{(b)} Based on Intensity Variance}
	\label{fig:functional_compare}
\end{figure}

From the entire results, the data was grouped into different types of paint categories such as LiDAR functionalised, non-LiDAR functionalised, metallic, non-metallic, glossy and matte for easier comparison. This data was organised into bar charts and after further analysis, some observations are highlighted.

The graphs in Fig. \ref{fig:functional_compare} show the intensity average and intensity variance when separated into LiDAR functionalised and non-LiDAR functionalised paint type categories for comparison. A few observations were made based on the bar graphs. Glossy paints have very similar effects on LiDAR performance regardless of whether it is LiDAR functionalised or not by looking at both Fig. \ref{fig:ave_functional} \& \ref{fig:var_functional}. However, for matte paints, having LiDAR functionalised properties does significantly increase the reflected intensity the LiDAR records as compared to not having LiDAR functionalised properties. This is observed in Fig. \ref{fig:ave_functional}.

\begin{figure}[h!]
    \centering
	\captionsetup{justification=centering}
	\subfloat[]{\label{fig:ave_metallic} \includegraphics[width=0.48\columnwidth]{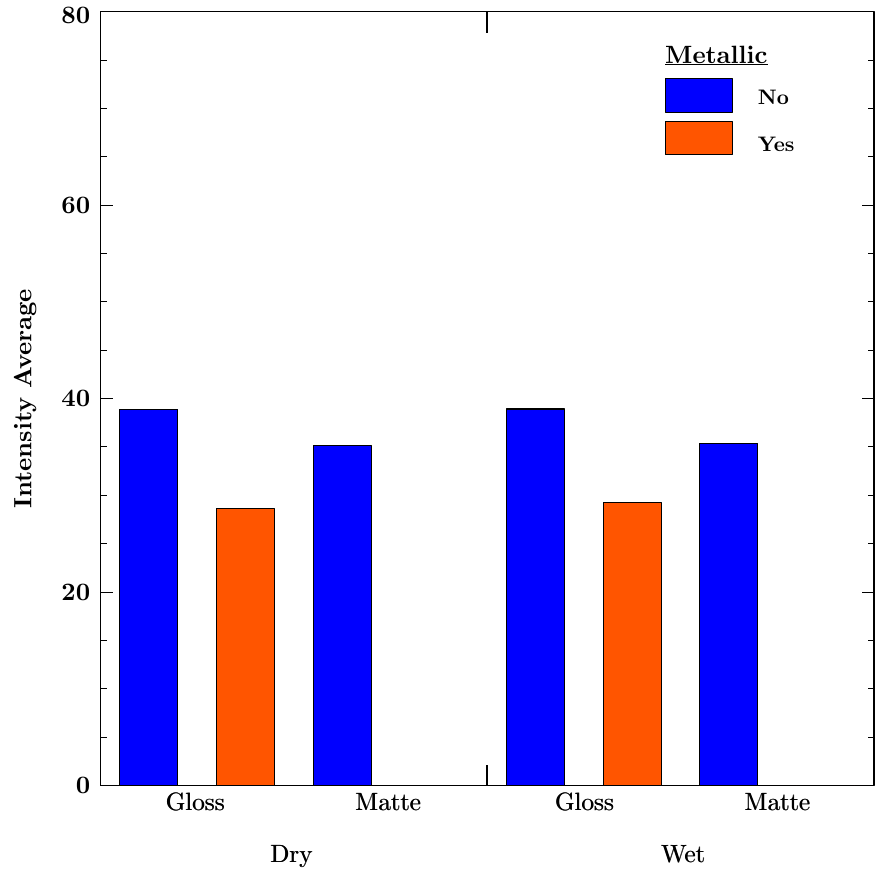}}
	\hspace{5pt}
	\subfloat[]{\label{fig:var_metallic} \includegraphics[width=0.48\columnwidth]{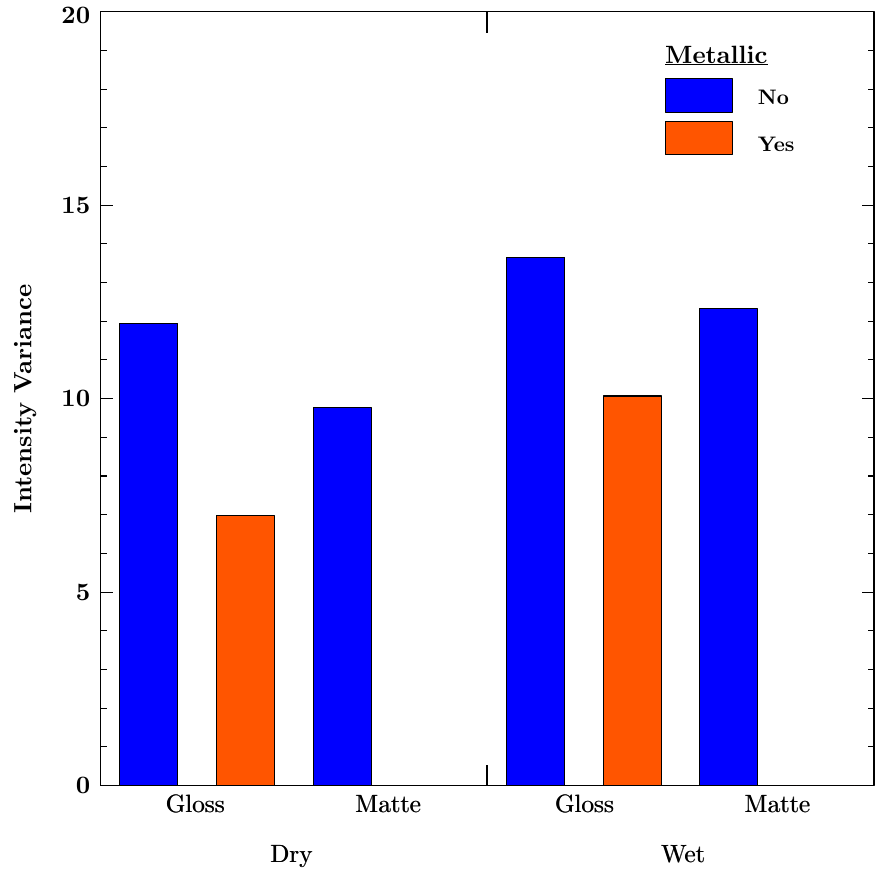}}
	\caption{Metallic paint type comparison \textbf{(a)} Based on Intensity Average  \textbf{(b)} Based on Intensity Variance}
	\label{fig:metallic_compare}
\end{figure}

The graphs in Fig. \ref{fig:metallic_compare} show the intensity average and intensity variance when separated into metallic and metallic paint type categories for comparison. Some interesting observations were made based on the bar graphs. Non-metallic glossy paints tend to show higher average intensity than metallic glossy paints which show lower average intensity in all conditions as seen in Fig. \ref{fig:ave_metallic}. The difference is also true when the paints are compared based on intensity variance. Non-metallic glossy paints tend to show higher variance intensity than metallic glossy paints which show lower variance intensity in all conditions as seen in Fig. \ref{fig:var_metallic}. Unfortunately, such a comparison cannot be made for matte paints because there are no metallic matte paint panels provided by NIPSEA.

\subsubsection{Tests with different target angles}
\begin{figure}[h!]
    \centering
	\captionsetup{justification=centering}
	\subfloat[]{\label{fig:ave_angles_dry} \includegraphics[width=0.48\columnwidth]{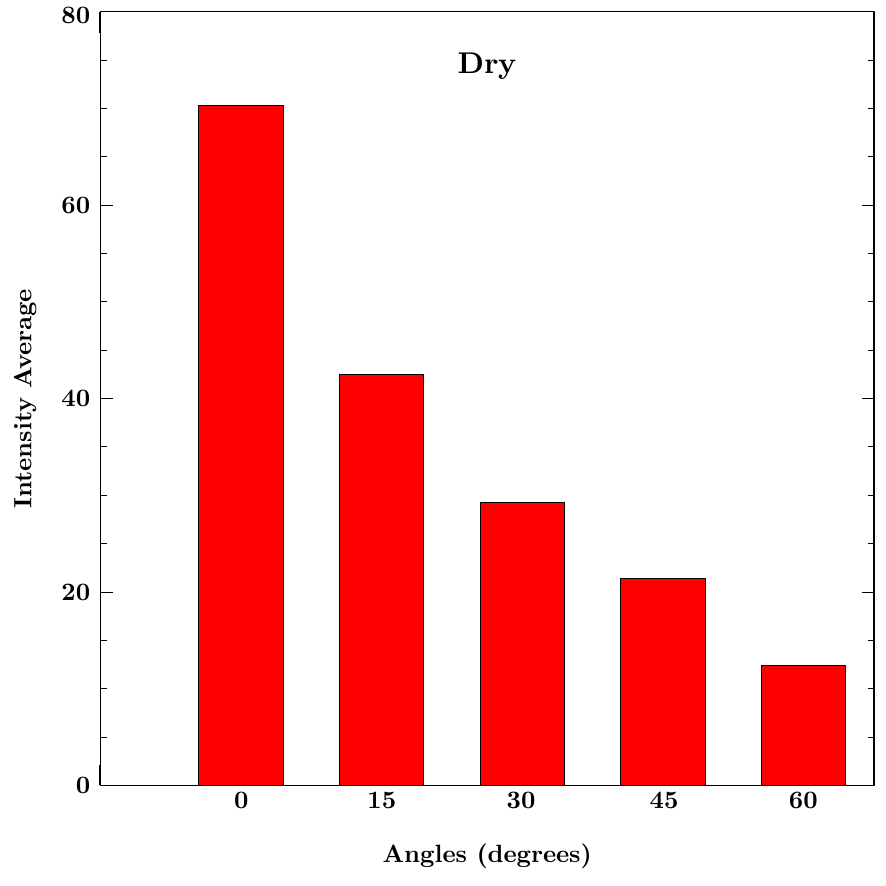}}
	\hspace{5pt}
	\subfloat[]{\label{fig:ave_angles_wet} \includegraphics[width=0.48\columnwidth]{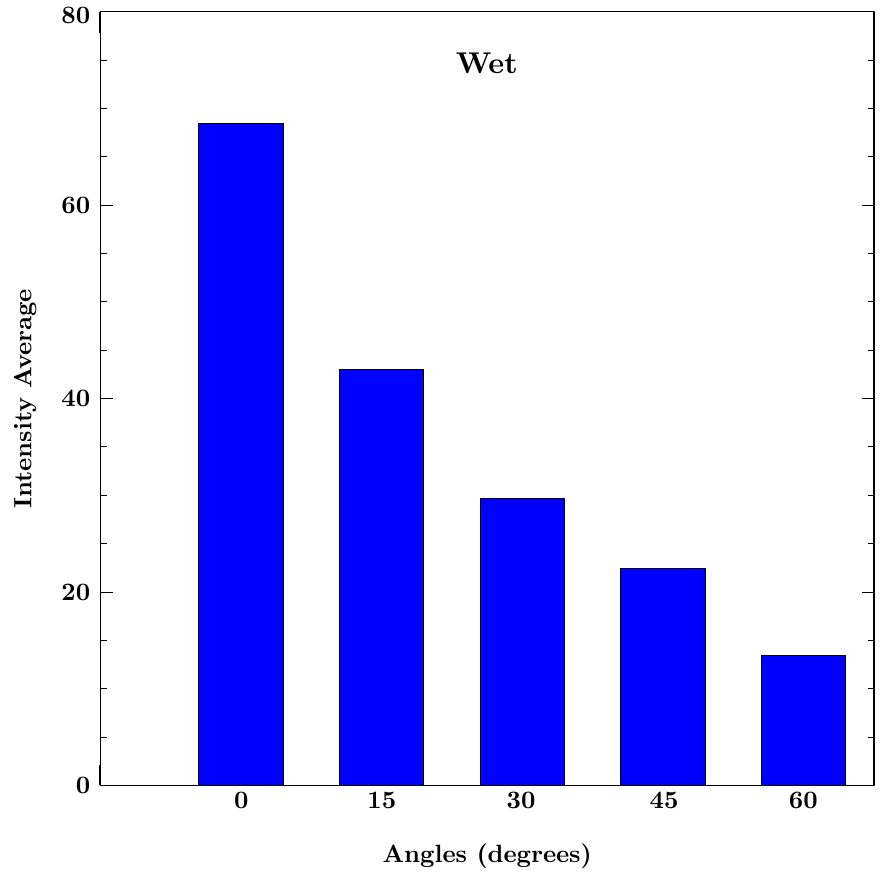}}
	\caption{Average reflected intensity vs. Target angles across all distances \textbf{(a)} Dry  \textbf{(b)} Wet}
	\label{fig:ave_angles}
\end{figure}

The tests with different target angles was done to emulate the effects of various environmental object orientations on a LiDAR. Five different angle orientations (0\degree, 15\degree, 30\degree, 45\degree, 60\degree) of the test targets were chosen. The results of these tests are illustrated in Fig. \ref{fig:ave_angles}.

The results clearly show that angle of the test targets has an impact on LiDAR performance. As the target angles increases, the average intensity decreases for both dry (shown in Fig. \ref{fig:ave_angles_dry}) and wet (shown in Fig. \ref{fig:ave_angles_wet}) surface conditions. Thus, this helps to ascertain that target object’s surface angles has an effect on LiDAR data outputs.

\subsubsection{Tests with different target distances}
\begin{figure}[h!]
    \centering
	\captionsetup{justification=centering}
	\subfloat[]{\label{fig:ave_dist_dry} \includegraphics[width=0.48\columnwidth]{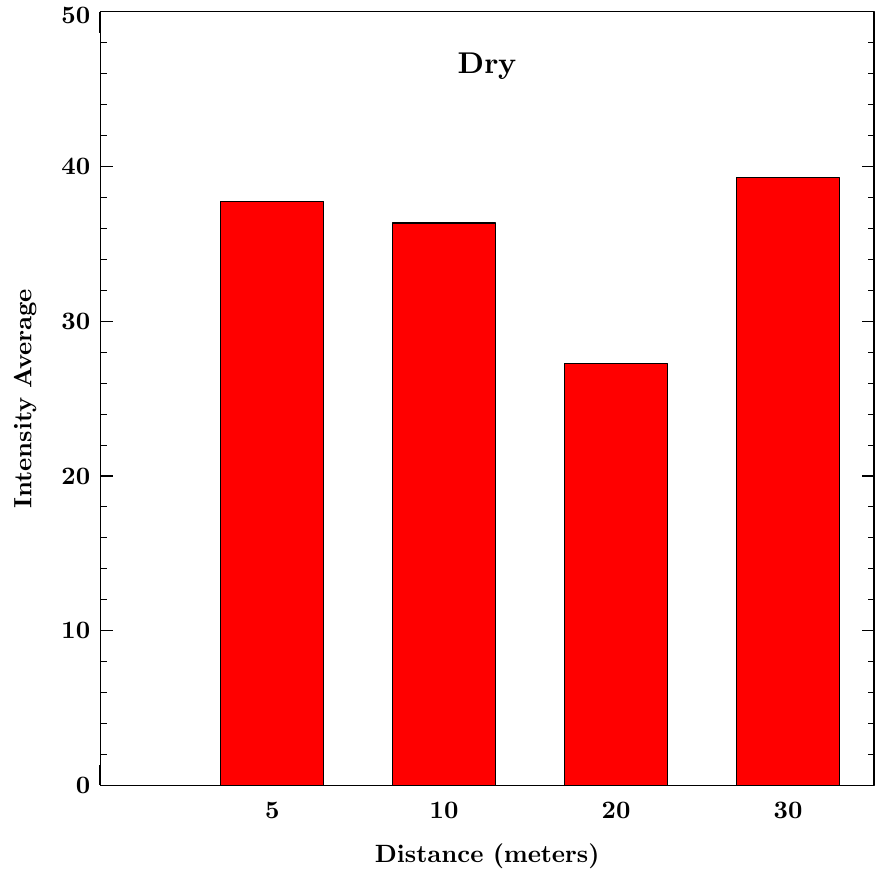}}
	\hspace{5pt}
	\subfloat[]{\label{fig:ave_dist_wet} \includegraphics[width=0.48\columnwidth]{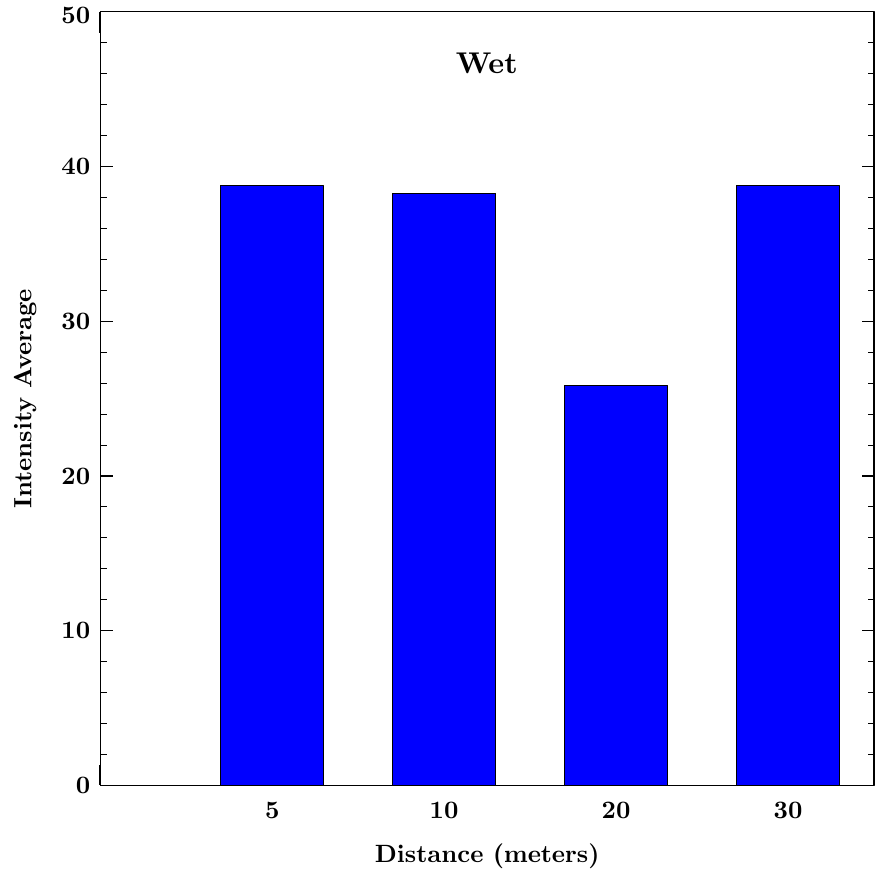}}
	\caption{Average reflected intensity vs. Target distances across all angles \textbf{(a)} Dry  \textbf{(b)} Wet}
	\label{fig:ave_dist}
\end{figure}

The tests with different target distances was performed to mimic the effects of various environmental object distances on a LiDAR. Four target distances (5.0m, 10.0m, 20.0m, 30.0m) were selected for the tests. The results for these tests are displayed in Fig. \ref{fig:ave_dist}.

The above graphs show that target distances do affect LiDAR performance to a substantial extent. Focusing on the 5m to 20m data for both dry (shown in Fig. \ref{fig:ave_dist_dry}) and wet (shown in Fig. \ref{fig:ave_dist_wet}) conditions, there is a downward trend for the average intensity values as the distance of the target to the LiDAR increases. A sharp drop in average intensity values is also noticed between the distances of 10m and 20m. Looking only at the data collected for the distance of 30m for both dry and wet conditions, it is largely unreliable as the number of LiDAR points on the test targets at that distance is very low, making the data highly susceptible to disturbances from the surroundings. Hence, by analysing results of the Average reflected intensity vs. Target distances graphs, a target object’s distance from a LiDAR has an effect on LiDAR data outputs.

\subsubsection{Tests with different target surface conditions}
\begin{figure}[h!]
    \centering
	\captionsetup{justification=centering}
	\subfloat[]{\label{fig:Surface_cond_ave} \includegraphics[width=0.48\columnwidth]{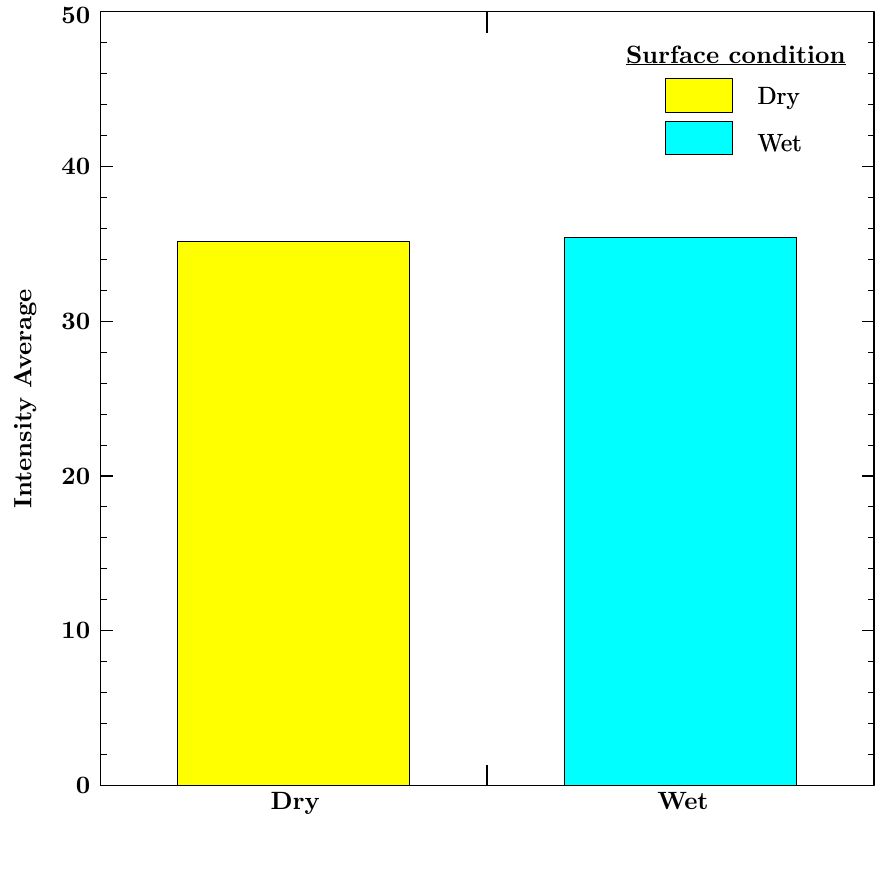}}
	\hspace{5pt}
	\subfloat[]{\label{fig:Surface_cond_var} \includegraphics[width=0.48\columnwidth]{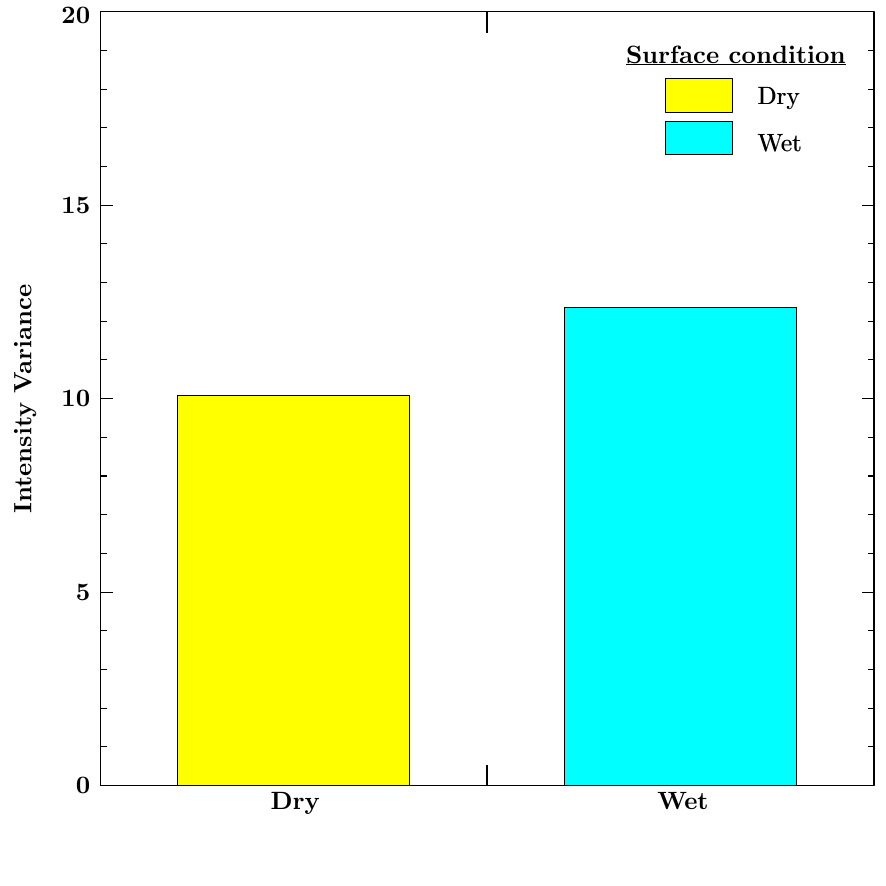}}
	\caption{\textbf{(a)} Intensity average  \textbf{(b)} Intensity variance for different surface conditions}
	\label{fig:Surface_cond}
\end{figure}

The tests with different target surface conditions were carried out to measure the effects of dry and wet surface conditions on LiDAR performance. For dry conditions, the test targets were just tested at the distances and angles as stated in the earlier chapter. For wet conditions, each of the test targets was first sprayed with water at 0\degree ~angle and 5m, with no respraying of water at other angles (15\degree, 30\degree, 45\degree, 60\degree). At each further distance, there was also spraying of water at 0\degree, but not for other angles. LiDAR data recording was started around 1 second after each manual spraying of water to allow the water droplets to settle on the test target's surface. These data recorded for both dry and wet surface conditions are shown in Fig. \ref{fig:Surface_cond}.

The data in the graph below shows that surface conditions have negligible effects on the average reflected intensity a LiDAR receives. However, for wet surfaces there is a slightly larger variation in the intensity values as seen from the higher intensity variance for wet surfaces as compared to a dry surface.

\subsection{Simulation}
\begin{figure}[h!]
    \centering
    \centerline{
        \includegraphics[width=.9\textwidth]{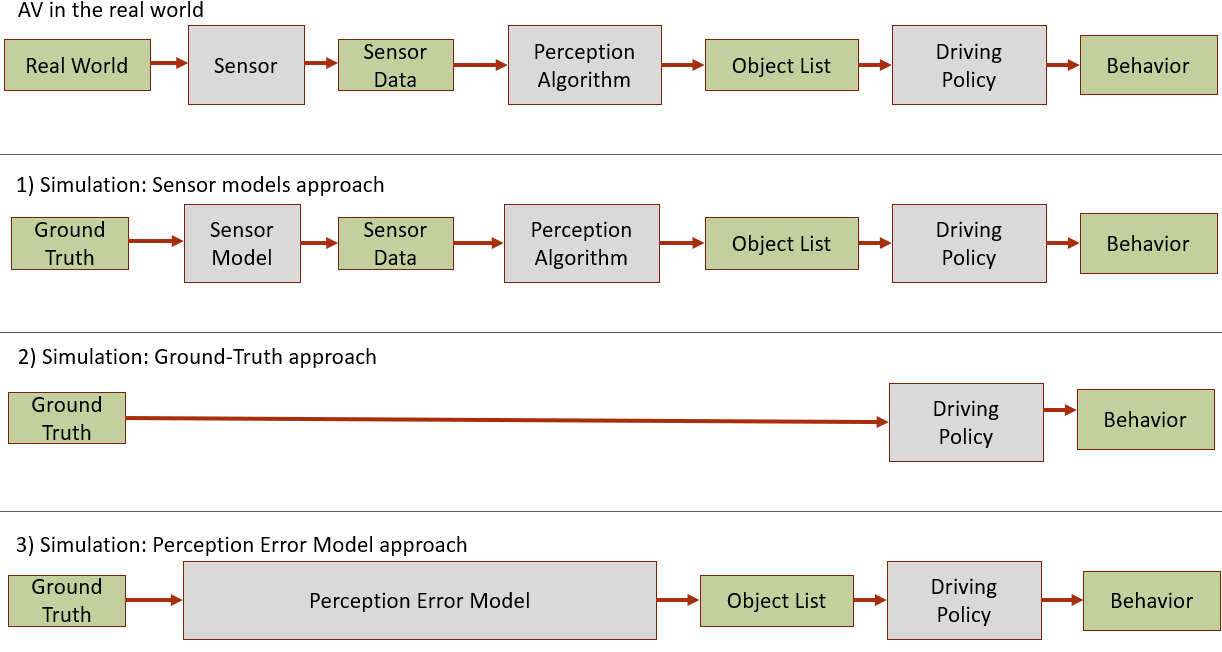}
    }
    \caption{Abstraction of a typical AV architecture a real-world deployment,  (1) simulation adopting sensor models, (2) ground-truth based simulation, (3) simulation with perception errors models.}
    \label{fig:simPerc}
\end{figure}

Analyzing the effect of various environmental parameters is helpful in the integration of sensing and perception modules into a simulation pipeline.
In \autoref{fig:simPerc}, we abstract a typical AV architecture and compare it with 3 different approaches for AV simulations.  
The first diagram in \autoref{fig:simPerc} depicts an abstraction of a typical AV architecture deployed in the real world.
The most direct adaptation of such architecture for virtual testing is to adopt sensor models and integrate the perception algorithm as is in the pipeline. 
Alternatively, the second approach could completely omit the sensing and perception module, leading to a simulation based on ground truth.
The last approach employs Perception Error Models (PEMs) \cite{piazzoni2020modeling,piazzoni2023simulation} to directly inject perception errors on the ground truth, generating an object list that is not perfect like the ground truth.

By comparing the three different simulation approaches, it is evident that the second approach does not contemplate any environmental conditions or perception errors. Still, this setup is the simplest to implement and, albeit unrealistic, affords studies on AV behavior under perfect perception. 

For a more complete and meaningful simulation, however, the effects of environmental parameters must be accounted for.
Adopting sensor models or PEMs leads to slightly different architectures, providing additional side benefits for AV development.
In particular, there are these major differences:
\begin{itemize}
    \item \textbf{Integration of environmental parameters}: In the case of sensor models, the environmental parameters will directly affect the sensor data. In the case of LiDAR, by looking at \autoref{fig:ave_angles} we can see how an effective sensor model should account for the angle of incidence of the LiDAR beam and tune the intensity of the point accordingly. 
    In the case of PEMs, instead, the point cloud is not generated. However, since we determined that the angle information leads to different errors in the synthetic data, we must assume that a perception algorithm may (or may not) provides objects with a different degree of error if the underlying sensor data is also affected by different errors. Thus, a PEM may also need to account for the environmental parameter when generating the object list.
    \item \textbf{Perception module}: PEMs are calibrated with the inclusion of the perception module. Thus, any change and alteration to this module would require a recalibration of the PEM. On the other hand, sensor models are more flexible and do not need recalibration if the perception module changes.
    \item \textbf{Computational Cost}: Sensor models must generate synthetic sensor data during the simulation loop. This task is significantly costly in terms of computational power. It scales poorly with the amount of sensors (as each sensor is modeled individually) and the quality of the sensors themselves. For example, increasing the image resolution in a camera or the number of layers in a LiDAR, directly increases the amount of synthetic data the model needs to generate during the simulation.
    On the other hand, PEMs are more efficient. First, they operate on the object level, where the information amount is significantly smaller than the sensor data. Secondly, their complexity does not directly increase  with the number of sensors and their quality, as the sensor and perception module is modeled as a whole. Lastly, there is no requirement to run the perception module of the AV during the simulation loop, which is one of the major computational costs of the AV stack.
    \item \textbf{Other benefits}: The capability of generating realistic synthetic data can benefit other tasks, such as training the perception module. However, the effectiveness of augmenting the training data with synthetic data is still being explored. In particular, one major shortcoming is due to the actual fidelity of the synthetic data, which is often not comparable to real data. 
    PEMs, on the other hand, also provide a more interpretable understanding of the perception errors affecting the AV. Moreover, it allows  direct control of the errors injected in the simulation since they can be easily tuned.
\end{itemize}

To summarize, a PEM-based simulation is a compromise between a simulation based on ground truth (simple but very unrealistic) and one based on sensor models (more realistic but complex and, thus, costly).

\subsubsection{Perception Error Model Requirements}
As seen in \autoref{fig:simPerc}, a PEM has the role to read the ground truth from the simulator generate the Object List for the AV. 
Intuitively, the objects list can be seen as the AV \textit{perception} of the ground truth. 
In particular, the ground truth will describe the parameters and properties of all the surroundings object in the simulation, and the object list describe the AV internal representation of the surroundings.
This observation implies that the ground truth and the object list are, in abstract, describing the same data. 
It follows that a PEM has to model the differences (i.e., errors) that affect the object list.
There are mainly 2 types of errors that can affect the object list:
\begin{itemize}
    \item Detection: the object list may be have false positive and false negatives (missing objects).
    \item Object Parameters: each parameter of any object has the potential of be affected by errors. For example, the position of an object may not be measured accurately, and thus may appear with wrong values in the object list.
\end{itemize}
    
Thus, to implement a PEM is sufficient to compare ground truth data and the respective object list data \cite{piazzoni2023simulation}.
In fact, by computing their difference (i.e., missing objects, parameter values difference) we can obtain a dataset of the perception errors. 
This dataset can the analysed via statistical and machine learning/deep learning towards obtaining a model of such errors.
This procedure is analogue to the procedure of evaluating the perception module performance via standard computer vision metrics. The main difference is that instead of trying to summarize the perception module performance as much as possible to facilitate comparison, a PEM aims to capture the varying conditions causing alterations in such performance.

\section{Conclusion}
This white paper provides an overview of the effects of selected environmental parameters on AV sensors. It presents our experiments to verify and validate the impact of these parameters on a LiDAR sensor through physical testing. From the results and analysis of the tests, we have gained some insights on which environmental parameters have the greatest effect on LiDAR visibility and performance. Our results show that the type of paint on the target object, the angle orientation of the target object and the distance of the target object all have large impacts on LiDAR performance. While target surface conditions have a smaller impact on LiDAR visibility.
Thus, they should be considered as conditioning effects on the error models at both the sensor data and objects list level.

Furthermore, we have better understood AV sensors' weaknesses and challenges, particularly for LiDAR sensors. This knowledge will help to provide useful information for any future testing of AV sensors by the various stakeholders in the AV industry, whether physically or virtually. We see the potential of utilizing the results obtained from this study to advance our research into producing virtual simplified sensor models with the errors and weaknesses that AV sensors face. This enables simulators to accurately represent imperfections of real-world sensors, allowing for more realistic modeling of AV sensing and perception systems virtually.  

\subsection{Future works}
This study explores a relatively limited amount of the possible combinations of environmental conditions and sensors. In particular, different materials such as cloth, plastic, metal, glass, or concrete could be explored in a similar study. 
Moreover, evolving sensor technologies and alternative solutions may provide different results.
Thus, future works should explore different environmental conditions, materials, and sensors.

On the simulation side, Perception Errors Models are still an under-explored field. In particular, their integration in simulator is not standardized, and their development and validation process may require ad-hoc solution for each AV (i.e., sensor setup and perception module).
Moreover, challenges in data collection and labeling lead to a limited amount of data for the training and validation process of both sensor models and PEMs.

\section{Acknowledgements}
\label{sec:Acknowledgements}

We acknowledge our current partners who are supporting us in this project. In particular, we thank NIPSEA Technologies Pte. Ltd. for providing us with a diverse set of automotive paint panels and the ongoing collaboration.

This research/project is supported by the National Research Foundation, Singapore, and Land Transport Authority under Urban Mobility Grand Challenge (UMGC-L010). Any opinions, findings and conclusions or recommendations expressed in this material are those of the author(s) and do not reflect the views of National Research Foundation, Singapore, and Land Transport Authority.

\cleardoublepage

\printbibliography 

@misc{standard2018j3016,
  title="{J3016\_201806 standard: Taxonomy and Definitions for Terms Related to Driving Automation Systems for On-Road Motor Vehicles}",
  author={SAE},
  year={2018}
}

@misc{Apollo_perception,
author={Apollo},
title = {{Apollo Perception, \url{https://github.com/ApolloAuto/apollo/tree/master/modules/perception }}},
year={2021}
}

@misc{umgc_whitepaper1,
      title={White paper on LiDAR performance against selected Automotive Paints}, 
      author={James Lee Wei Shung and Paul Hibbard and Roshan Vijay and Lincoln Ang Hon Kin and Niels de Boer},
      year={2023},
      eprint={2309.01346},
      archivePrefix={arXiv},
      primaryClass={cs.RO}
}

@article{Lambert2020,
author = {Lambert, Jacob and Carballo, Alexander and Cano, Abraham Monrroy and Narksri, Patiphon and Wong, David and Takeuchi, Eijiro and Takeda, Kazuya},
doi = {10.1109/ACCESS.2020.3009680},
issn = {21693536},
journal = {IEEE Access},
pages = {131699--131722},
title = {{Performance Analysis of 10 Models of 3D LiDARs for Automated Driving}},
volume = {8},
year = {2020}
}

@article{Beiker2018,
author = {Beiker, Sven},
pages = {36},
title = {{Unsettled Topics Concerning Sensors for Automated Road Vehicles}},
year = {2018}
}

@article{marti2019review,
  title={A review of sensor technologies for perception in automated driving},
  author={Marti, Enrique and de Miguel, Miguel Angel and Garcia, Fernando and Perez, Joshue},
  journal={IEEE Intelligent Transportation Systems Magazine},
  volume={11},
  number={4},
  pages={94--108},
  year={2019},
  publisher={IEEE}
}

@article{li2020lidar,
  title={Lidar for autonomous driving: The principles, challenges, and trends for automotive lidar and perception systems},
  author={Li, You and Ibanez-Guzman, Javier},
  journal={IEEE Signal Processing Magazine},
  volume={37},
  number={4},
  pages={50--61},
  year={2020},
  publisher={IEEE}
}

@misc{Infographic,
author={Hokuyo},
title = {{INFOGRAPHIC: 2D vs 3D LiDAR Sensors, \url{https://hokuyo-usa.com/resources/blog/2d-vs-3d-lidar-sensors }}},
year={2021}
}

@article{van2018autonomous,
  title={Autonomous vehicle perception: The technology of today and tomorrow},
  author={Van Brummelen, Jessica and O’Brien, Marie and Gruyer, Dominique and Najjaran, Homayoun},
  journal={Transportation research part C: emerging technologies},
  volume={89},
  pages={384--406},
  year={2018},
  publisher={Elsevier}
}

@misc{whylidar,
author={Leddartech},
title = {{Why LiDAR,
\url{https://leddartech.com/why-lidar }}},
year={Retrieved on 19/01/2022}
}

@inproceedings{wang2018characterization,
  title={Characterization of a RS-LiDAR for 3D Perception},
  author={Wang, Zhe and Liu, Yang and Liao, Qinghai and Ye, Haoyang and Liu, Ming and Wang, Lujia},
  booktitle={2018 IEEE 8th Annual International Conference on CYBER Technology in Automation, Control, and Intelligent Systems (CYBER)},
  pages={564--569},
  year={2018},
  organization={IEEE}
}

@inproceedings{pomerleau2012noise,
  title={Noise characterization of depth sensors for surface inspections},
  author={Pomerleau, Fran{\c{c}}ois and Breitenmoser, Andreas and Liu, Ming and Colas, Francis and Siegwart, Roland},
  booktitle={2012 2nd International Conference on Applied Robotics for the Power Industry (CARPI)},
  pages={16--21},
  year={2012},
  organization={IEEE}
}

@article{rasshofer2011influences,
  title={Influences of weather phenomena on automotive laser radar systems},
  author={Rasshofer, Ralph H and Spies, M and Spies, H},
  journal={Advances in Radio Science},
  volume={9},
  number={B. 2},
  pages={49--60},
  year={2011},
  publisher={Copernicus GmbH}
}

@article{carrea2016correction,
  title={Correction of terrestrial LiDAR intensity channel using Oren--Nayar reflectance model: An application to lithological differentiation},
  author={Carrea, Dario and Abellan, Antonio and Humair, Florian and Matasci, Battista and Derron, Marc-Henri and Jaboyedoff, Michel},
  journal={ISPRS Journal of Photogrammetry and Remote Sensing},
  volume={113},
  pages={17--29},
  year={2016},
  publisher={Elsevier}
}

@misc{reflectivityofpaint,
author={Christopher M. Seubert},
title = {{IR Reflectivity of Paint:
Autonomy and CO2 Emissions,
\url{https://detroitcc.org/wp-content/uploads/2018/07/IR-Reflectivity-of-Paint-Autonomy-and-CO2-Seubert.pdf }}},
year={Retrieved on 21/01/2022}
}

@misc{vlp16_usermanual,
author={Velodyne},
title = {{Velodyne VLP-16 User Manual,
\url{https://velodynelidar.com/wp-content/uploads/2019/12/63-9243-Rev-E-VLP-16-User-Manual.pdf}}},
year={Retrieved on 21/01/2022}
}

@article{rosique2019systematic,
  title={A systematic review of perception system and simulators for autonomous vehicles research},
  author={Rosique, Francisca and Navarro, Pedro J and Fern{\'a}ndez, Carlos and Padilla, Antonio},
  journal={Sensors},
  volume={19},
  number={3},
  pages={648},
  year={2019},
  publisher={Multidisciplinary Digital Publishing Institute}
}

@misc{NIST,
  author = {Geraldine Cheok and Stefan Leigh and Andrew Rukhin},
  title = {Calibration Experiments of a Laser Scanner},
  year = {2002},
  month = {2002-09-01},
  publisher = {NIST Interagency/Internal Report (NISTIR), National Institute of Standards and Technology, Gaithersburg, MD},
  url = {https://tsapps.nist.gov/publication/get_pdf.cfm?pub_id=860455},
  language = {en},
}

@inproceedings{raju2019_autoware_apollo,
  title={Performance of Open Autonomous Vehicle Platforms: Autoware and Apollo},
  author={Raju, Vysyaraju Manikanta and Gupta, Vrinda and Lomate, Shailesh},
  booktitle={2019 IEEE 5th International Conference for Convergence in Technology (I2CT)},
  pages={1--5},
  year={2019},
  organization={IEEE}
}

@book{toda2012automotive,
  title={Automotive painting technology: A Monozukuri-Hitozukuri perspective},
  author={Toda, Kimio and Salazar, Abraham and Saito, Kozo},
  year={2012},
  publisher={Springer}
}

@book{automotive_handbk,
  title={Automotive paint handbook: paint technology for auto enthusiasts \& body shop professionals},
  author={Pfanstiehl, John},
  year={1998},
  publisher={Penguin}
}

@article{behroozpour2017lidar,
  title={Lidar system architectures and circuits},
  author={Behroozpour, Behnam and Sandborn, Phillip AM and Wu, Ming C and Boser, Bernhard E},
  journal={IEEE Communications Magazine},
  volume={55},
  number={10},
  pages={135--142},
  year={2017},
  publisher={IEEE}
}

@inproceedings{anand2022evaluation,
  title={Evaluation of the quality of LiDAR data in the varying ambient light},
  author={Anand, Bhaskar and Verma, Harshal and Thakur, Abhishek and Alam, Parvez and Rajalakshmi, P},
  booktitle={2022 IEEE Sensors Applications Symposium (SAS)},
  pages={1--5},
  year={2022},
  organization={IEEE}
}

@inproceedings{xique2018evaluating,
  title={Evaluating complementary strengths and weaknesses of ADAS sensors},
  author={Xique, Ismael J and Buller, William and Fard, Zahra Bahrani and Dennis, Eric and Hart, Benjamin},
  booktitle={2018 IEEE 88th Vehicular Technology Conference (VTC-Fall)},
  pages={1--5},
  year={2018},
  organization={IEEE}
}

@inproceedings{piazzoni2020modeling,
  title     = {Modeling Perception Errors towards Robust Decision Making in Autonomous Vehicles},
  author    = {Piazzoni, Andrea and Cherian, Jim and Slavik, Martin and Dauwels, Justin},
  booktitle = {Proc. of the 29th International Joint Conference on
               Artificial Intelligence, {IJCAI-20}},
  pages     = {3494--3500},
  year      = {2020},
  month     = {7},
  note      = {Main track}
}

@article{piazzoni2023simulation,
  title={On the Simulation of Perception Errors in Autonomous Vehicles},
  author={Piazzoni, Andrea and Cherian, Jim and Dauwels, Justin and Chau, Lap-Pui},
  journal={arXiv preprint arXiv:2302.11919},
  year={2023}
}

\clearpage

\renewcommand{\thefigure}{\thesection.\arabic{figure}}  
\setcounter{figure}{0} 
\renewcommand{\thetable}{\thesection.\arabic{table}}    
\setcounter{table}{0} 

\end{document}